\documentclass[journal]{new-aiaa}

\usepackage[utf8]{inputenc}
\usepackage{graphicx}

\usepackage{amssymb}
\usepackage{amssymb}
\usepackage{textcomp}
\usepackage[version=4]{mhchem}
\usepackage{siunitx}
\usepackage{longtable,tabularx}
\usepackage{booktabs}
\usepackage{subcaption}
\usepackage{float}
\usepackage{hyperref}
\usepackage{tikz}
\usetikzlibrary{positioning,arrows.meta,shapes.geometric,fit,calc,patterns,matrix,decorations.pathreplacing}
\newcommand{\xobs}{\mathbf{X}^{\text{obs}}}
\newcommand{\xfut}{\mathbf{X}^{\text{fut}}}
\newcommand{\Rbb}{\mathbb{R}}

\tikzset{
  block/.style={rectangle, draw, rounded corners=2pt, fill=#1!15,
                minimum height=0.7cm, minimum width=1.8cm, align=center,
                font=\small},
  block/.default=blue,
  arrow/.style={-{Stealth[length=4pt]}, thick},
  darrow/.style={<->,thick,gray},
}

\title{FlowATC: Aircraft Trajectory Prediction via Flow Matching}

\author{Mathurin Petit\footnote{Graduate Student, \'Ecole Polytechnique; e-mail: \href{mailto:mathurin.petit.23@polytechnique.org}{mathurin.petit.23@polytechnique.org} (corresponding author).}}
\affil{\'{E}cole Polytechnique, Palaiseau, 91128, France}

\author{Emir Torun\footnote{Undergraduate Student, Technische Universität Berlin, Berlin, Germany.}}
\affil{Technische Universität Berlin, Berlin, 10623, Germany}

\author{Louis Brusset\footnote{Graduate Student, Graduate Mines Paris, PSL University}}
\affil{Mines Paris--PSL University, Paris, 75006, France}

\author{Jordan Kam\footnote{Ph.D. Student, Department of Aerospace Engineering, California Institute of Technology, AIAA Student Member.}}
\affil{California Institute of Technology, Pasadena, CA, 91125, USA}

\author{ Alexandre M.\ Bayen\footnote{Professor, Department of Electrical Engineering and Computer Sciences,
  University of California, Berkeley.}}
\affil{University of California, Berkeley, Berkeley, CA 94720, USA}

\begin{document}

\maketitle

\begin{abstract}
Building accurate decision-support tools
for next generation air traffic control requires robust trajectory prediction models. We present a flow-matching architecture trained exclusively on historical aircraft trajectories, with no route labels or chart supervision.
Trained on 1.15 million Automatic Dependent Surveillance--Broadcast trajectory windows collected over the San Francisco Bay Area, the model generates aircraft trajectory distributions that closely match historical traffic, reproducing known airspace structure around San Francisco Airport such as the shape of SFO's published NIITE FOUR departure procedure. Our model is trained directly on the native, irregular ADS-B sampling interval. Trajectory prediction is cast as sequence inpainting using a block-causal Transformer that denoises future state tokens conditioned on the observed history using Conditional Flow Matching or Denoising Diffusion Probabilistic Models. We compare our architecture against constant-velocity, deterministic-Long Short Term Memory, and Conditional Variational Autoencoders baselines. At matched parameter count, CFM outperforms DDPM by 11--26\% in minADE@20, and both generative objectives surpass the CVAE baseline by 31--41\%. We further show that the error degrades gracefully with prediction horizon, and the architecture remains effective when retrained on temporally decimated feeds. Lastly, we sample $K$ independent completions, yielding spatial probabilistic occupancy estimates that can serve as input to downstream conflict-risk estimation.
\end{abstract}

\section*{Nomenclature}

{\renewcommand\arraystretch{1.0}
\noindent\begin{longtable*}{@{}l @{\quad=\quad} l@{}}
$\mathbf{X}$          & trajectory window, $T \times F$ array of ADS-B samples \\
$\xobs$               & observation prefix, first $T_{\text{obs}}$ steps \\
$\xfut$               & future suffix, next $T_{\text{fut}}$ steps to predict \\
$T$                   & total sequence length \\
$F$                   & features per timestep  \\
$\phi,\lambda$        & geodetic latitude and longitude \\
$d$                   & Transformer hidden dimension \\
$u_\theta$            & CFM velocity field network \\
$p_i = (x_i, y_i)$ & Lateral position at trajectory timestep $i$ \\
$K$                   & number of samples in best-of-$K$ evaluation \\
$\text{minADE@}K$     & min-over-$K$ average displacement error, m \\
$\text{minFDE@}K$     & min-over-$K$ final displacement error, m \\
$\text{NLL@}n$        & KDE negative log-likelihood at future step $n$ \\
\end{longtable*}}
\addtocounter{table}{-1}

\section{Introduction}

\lettrine{L}{o}w altitude \textit{air traffic control} (ATC) relies on accurate short-horizon aircraft trajectory prediction to maintain safe separation.
\textit{Automatic Dependent Surveillance--Broadcast} (ADS-B) data provides continuous position updates across the \textit{National Airspace System} (NAS), yet forecasting where an aircraft will be in the next two minutes remains hard. A flight on a standard \textit{San Francisco} (SFO) arrival may turn left or continue straight depending on runway assignment and traffic sequencing, both invisible from position data alone.
Prediction is therefore inherently multimodal, meaning a single observed history is consistent with several physically plausible futures (or modes) \cite{ivanovic2018, prutsch2026ascent, yang2025goodflight}. The quantity that matters operationally is not only where an aircraft will be, but how probable a conflict is at that location, which calls for a predictor that returns a full distribution over futures rather than a single point estimate \cite{salzmann2020trajectron, jiang2023motiondiffuser, figuet2026cfm}.

The generative-modelling toolkit powering recent advances in image synthesis and robot motion planning transfers naturally to continuous trajectory data.
\textit{Denoising Diffusion Probabilistic Models} (DDPM)~\cite{ho2020ddpm} learn to reverse a Markov noising chain and, with the \textit{Denoising Diffusion Implicit Models} (DDIM) sampler~\cite{song2021ddim}, generate high-quality samples in a handful of steps.
\textit{Conditional Flow Matching} (CFM)~\cite{lipman2022flow} is a more direct alternative regressing a velocity field that transports Gaussian noise to data along straight-line paths, giving a simpler training objective and fewer integration steps. Both families produce samples rather than point estimates. Generating $K$ independent samples at inference yields a distribution over future positions that covers the inherently multimodal space of plausible aircraft states even when the model is conditioned on a single input modality. The Transformer architecture \cite{vaswani2017attention} represents data as a set of tokens coupled by self-attention. This tokenized view is attractive for trajectory prediction for two reasons: it is flexible (heterogeneous inputs, i.e., positions, time deltas, and, in the future, ATC voice or weather fields, are simply additional tokens) and fast on modern hardware. Crucially, the \textit{Diffusion Transformer} (DiT)~\cite{peebles2023dit} shows that a Transformer backbone, with the generation step injected through \textit{Adaptive Layer Normalization} (AdaLN), is an excellent denoiser for diffusion. Specifically, we exploit exactly this synergy: a DiT style backbone hooks up cleanly with both DDPM and CFM, letting us frame trajectory prediction as token-level inpainting.

\subsection*{Related work}

Classical predictors fall into at least four families.
\emph{Kinematic} models (e.g.\ constant velocity / constant turn) propagate the last observed state forward under a motion assumption; they are interpretable and fast but produce a single deterministic forecast which does not take into account surrounding airspace information. We use constant velocity as a lower bound as it provides a good estimate of the order of magnitude of the error for such prediction tasks.
\emph{Deterministic neural} models, typically an LSTM/Gated Recurrent Unit (GRU) encoder--decoder or a Transformer regressor~\cite{tong2023longterm,zhao2019lstm,Zeng2022Review}, learn data driven dynamics but still emit one trajectory per query, suppressing the uncertainty that matters for safety.
\emph{Direct multimodal} predictors instead decode a fixed set of plausible trajectory hypotheses. ASCENT~\cite{prutsch2026ascent}, for example, uses a Transformer encoder with learnable mode queries to predict multiple 3D future trajectories together with associated mode scores in non-towered terminal airspace, achieving strong best-of-$K$ performance on the TrajAir benchmark. Unlike stochastic generative models, however, such approaches represent multimodality through a finite set of explicitly decoded hypotheses rather than by sampling from a continuous conditional distribution.
\emph{Probabilistic generative} models output a distribution: \textit{Conditional Variational Autoencoders} (CVAE) such as Trajectron++~\cite{salzmann2020trajectron} are widely used probabilistic baselines for multimodal trajectory forecasting, sampling a latent variable to produce diverse futures. In autonomous driving, diffusion-based predictors have recently emerged as a strong alternative to CVAEs. By iteratively denoising noise samples to in distribution data, this technique offers a probabilistic approach to prediction tasks.  MotionDiffuser~\cite{jiang2023motiondiffuser} applies DDPM to multi-agent road-traffic forecasting and shows that diffusion captures multimodal distributions without trajectory anchors. Diffusion has also been applied directly to aircraft trajectory prediction: Yin et al.~\cite{yin2023aircraftdiffusion} combine the aircraft's history with contextual information representing intent and environmental conditions in a diffusion-based decoder, evaluated at Singapore Changi Airport, and GooDFlight~\cite{yang2025goodflight} first estimates goal positions, then generates diverse trajectories with a goal-guided diffusion decoder.
In the aerospace domain, Briden et al.~\cite{briden2025spacecraft} apply diffusion to spacecraft descent planning, framing trajectory solutions as composable probability density functions; our setting is complementary, targeting probabilistic prediction of civil aircraft from surveillance observations, where maneuver structure is governed by ATC procedures rather than road geometry. More recently, diffusion models have also been widely adopted for robot motion planning: \citet{janner2022diffuser} showed that full action trajectories can be generated by iterative denoising guided by reward functions.

Closest to the present work in generative formulation, Figuet et al.~\cite{figuet2026cfm} apply Conditional Flow Matching to short-term aircraft trajectory prediction, using a Transformer encoder--decoder pair. Their study targets en-route traffic above FL195 in Swiss Free Route Airspace, in an aircraft-centric frame normalized to the last observed state, with absolute position provided as an explicit 8-dimensional context vector rather than encoded in the trajectory features directly. ADS-B is resampled to a uniform 1\,Hz grid. Our study is complementary along four axes. First, we target low-altitude terminal airspace, where published procedures dictate maneuver structure and traffic mixes commercial and general aviation; we retain absolute Cartesian coordinates so this structure can be learned without chart supervision (Section~\ref{sec:spatial}). Second, we predict a $\approx$128\,s horizon at the native, irregular ADS-B sampling rate, rather than on a resampled grid. Third, we replace the encoder--decoder pair with a single DiT that ingests clean and noisy tokens by concatenation, casting the task as sequence inpainting. Fourth, we benchmark CFM against CVAE and DDPM at matched capacity across three model scales, isolating the contribution of the objective itself.

Our novel contributions include the following:

\begin{enumerate}[label=(\roman*)]
    \item A sequence-inpainting architecture concatenating observed and noisy tokens, block-causal self-attention, AdaLN time conditioning allowing a DiT to address trajectory completion without an intermediate encoder.
    \item Benchmarking constant-velocity, deterministic LSTM, and CVAE (Trajectron++) baselines against DDPM and CFM variants of the same backbone, and showing that CFM is substantially more accurate at equal parameter count.
    \item Characterizing operational flexibility: graceful degradation over the horizon, retraining on lower-rate streams, extrapolation beyond the training horizon, and inference cost.
    \item Showing that FlowATC recovers Bay Area airspace structure without chart supervision and that its output distribution is accurate: at airspace branch points it reproduces the distribution of maneuvers actually flown and gives each aircraft a close to calibrated distribution over its next maneuver.
\end{enumerate}

\section{Methodology}
\label{sec:data}

\begin{figure}[t]
    \centering
    \includegraphics[width=0.55\linewidth]{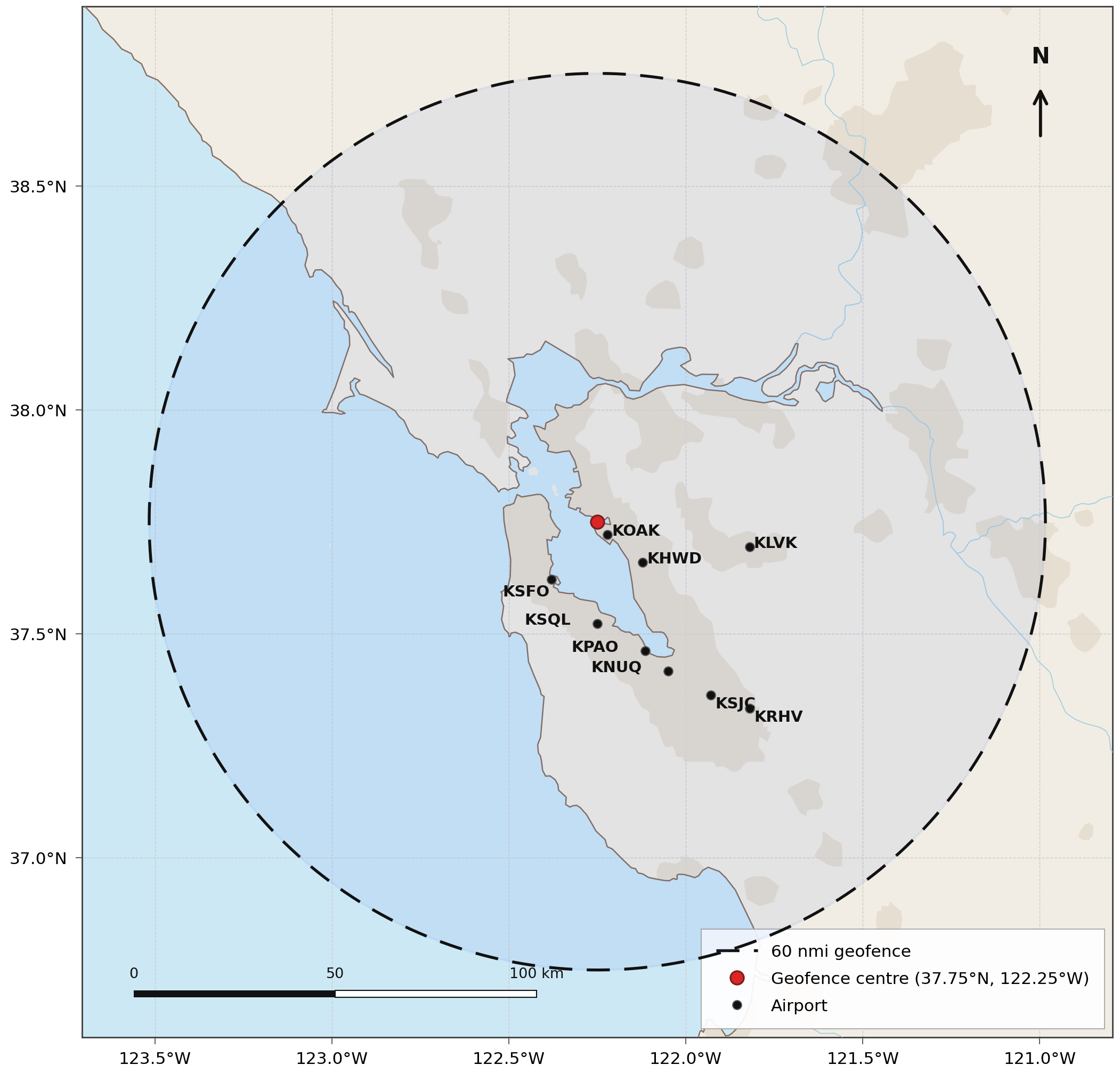}
    \caption{Geographic coverage of the ADS-B collection used for the present work. The dashed circle shows the $60$\,nautical-mile radius geofence centered on San Francisco Bay ($37.75^\circ$N, $122.25^\circ$W, red dot). Black dots mark the airports referenced in this study. The Cartesian coordinate frame used for trajectory representation (Eq.~\eqref{eq:cartesian}) is centered on SFO ($37.6213^\circ$N,$122.3790^\circ$W).}
    \label{fig:geo_coverage}
\end{figure}

\subsection{Aircraft Trajectory Data}

ADS-B is a cooperative surveillance technology in which an aircraft periodically broadcasts its own state: identity, position, altitude, velocity, and vertical rate, derived primarily from onboard Global Navigation Satellite System (GNSS). The broadcasts are received by a dense network of ground stations and can be aggregated by public feeds, making ADS-B a high-coverage, low-cost source of trajectory data over busy terminal airspace. We collect ADS-B continuously from the ADS-B LOL live feed (\href{https://adsb.lol/}{adsb.lol}) using the native API that queries aircraft states every $2\,s$ over a circular geofence of 60 nautical miles radius centered at $37.75^\circ$N, $122.25^\circ$W (San Francisco Bay). The scraper records, for each aircraft, the ICAO24 transponder code, callsign, Unix timestamp, longitude, latitude, barometric altitude, ground speed, true track, and vertical rate. Collection ran for 12 days, 10–22 April 2026, yielding 21{,}515{,}794 raw state vectors. The geofence and the airports referenced in this study are shown in Fig.~\ref{fig:geo_coverage}; dataset statistics are summarized in Table~\ref{tab:dataset}.

The collection spans a broad mix of Bay Area traffic (shares below by unique aircraft): large commercial aircraft (ICAO category A3, 41.8\%), light general aviation (A1, 32.1\%), heavy aircraft such as B747/A380 (A5, 10.8\%), and lighter traffic (remaining 15.3\%).

\begin{table}[bt!]
\caption{Bay Area ADS-B dataset statistics.}
\label{tab:dataset}
\centering
\small
\begin{tabular}{ll}
\toprule
Variable & Value \\
\midrule
Collection period              & 10 Apr–22 Apr 2026 (12 days) \\
Raw ADS-B state vectors        & 21,515,794 \\
Geographic coverage            & 60\,nm radius circle, centre 37.75°N, 122.25°W \\
Cartesian reference (SFO)      & 37.6213°N, 122.3790°W \\
Segmentation cut               & Gap $>$ 120\,s or callsign change \\
Features                       & $x,y,z,v_x,v_y,v_z$ \\
Sequence length / stride       & 86\,pts / 10\,pts \\
Total 86-point windows         & 1,349,388 \\
Training windows               & 1,149,245 \\
Validation windows             & 137,127 \\
Test windows                   & 63,016 \\
Intra-segment $\Delta t$: mean / median / std & 3.00\,s / 2.60\,s / 2.36\,s \\
\bottomrule
\end{tabular}
\end{table}

\subsection{Trajectory processing}

Raw ADS-B state vectors are segmented into continuous flight segments and windowed into 86-point sequences. Each window is split into a 43-point observed prefix and a 43-point future suffix to predict. The choice of a two-minute history and forecast horizon follows from discussions with pilots, who identified a 2-minute lookahead as the operationally relevant horizon. For maneuver prediction, each 43-point half spans roughly $128$\,s on average.

We project geodetic coordinates onto a local tangent-plane Cartesian frame centered on SFO, yielding a 6-dimensional feature vector $(x,y,z,v_x,v_y,v_z)$ entirely in meters or meters per second. Because absolute Cartesian coordinates encode geographic position, the model implicitly learns location-specific structure (e.g.\ that a south-westbound aircraft at $y\approx-15$\,km is on SFO final approach). Each feature is independently z-score normalized using training-split statistics.

Rather than resampling to a fixed grid, we expose the native, irregular ADS-B timing to the network directly via a learned time-delta embedding; full preprocessing details are given in Appendix~\ref{app:preprocessing}.

\section{Evaluation}

Predicting distributions over aircraft trajectories has until recently received moderate attention in the ATM literature~\cite{Zeng2022Review}, though generative formulations are now emerging~\cite{figuet2026cfm}. To account for the inherently multi-modal aspect of trajectories, encompassing the different acceptable maneuvers at some given point, this distributional point of view is necessary and is hard to measure in practice. Building upon Salzmann et al.'s Trajectron++\cite{salzmann2020trajectron}, we chose to evaluate our models on best of $K$ for \textit{Average/Final Displacement Error} (ADE/FDE) and \textit{Kernel Density Estimation Negative Log Likelihood} (KDE-NLL).

\subsection{Average and Final Displacement Error}

Let $\mathbf{p}_n = (x_n, y_n) \in \mathbb{R}^2$ denote the horizontal Cartesian position at future step $n$, extracted from ${\xfut}$. Because the variance on the vertical axis is secondary to horizontal variations, we chose to compute $z_n$ separately. Average displacement error accounts for how close the sampled trajectory is from the ground truth.
\begin{equation}
    \text{ADE}= \frac{1}{T_{\text{fut}}}\sum_{n} \|\hat{\mathbf{p}}_n - \mathbf{p}_n\|_2
\end{equation}

Note that this definition requires both predicted and ground-truth trajectories to share the same timestamps at each step $n$. In decimation experiments where the input sequence is downsampled, timestamp correspondence is preserved naturally. However, when predictions are requested at timestamps absent from the ground truth, the ground-truth positions $\mathbf{p}_n$ are obtained by linear interpolation to the desired evaluation points. To assess how far predictions drift at a fixed horizon, we report the Final Displacement Error, defined as the Euclidean distance between the predicted and ground-truth positions at the last future step $T_{\text{fut}}$:
\begin{equation}
    \text{FDE} = \|\hat{\mathbf{p}}_{T_{\text{fut}}} - \mathbf{p}_{T_{\text{fut}}}\|_2
\end{equation}
Because of the irregular ADS-B sampling, the final token is on average $128\,\text{s}$ in the future, with the central 80\% of windows spanning $[95\,\text{s},\, 165\,\text{s}]$, which equates to approximately one and a half to three minutes ahead.

\subsection{Best of K}
ADE and FDE alone are well suited for deterministic trajectory forecasting, where a single predicted trajectory is compared against the ground truth. In our setting, however, the future is genuinely multimodal: given the same observed prefix, an aircraft may initiate a left or right turn, continue en route, or enter a holding pattern, all equally valid outcomes invisible from position data alone. A deterministic metric would penalize any model that hedges across modes, even if one of its samples matches the ground truth closely. We therefore adopt the best-of-$K$ variants, minADE@$K$ and minFDE@$K$, standard in the multimodal forecasting literature~\cite{salzmann2020trajectron, jiang2023motiondiffuser}:

\begin{equation}
  \text{minADE@}K = \min_k \frac{1}{T_{\text{fut}}}\sum_{n} \|\hat{\mathbf{p}}^{(k)}_n - \mathbf{p}_n\|_2,
  \quad
  \text{minFDE@}K = \min_k \|\hat{\mathbf{p}}^{(k)}_{T_{\text{fut}}} - \mathbf{p}_{T_{\text{fut}}}\|_2 .
\end{equation}

These metrics reward sample coverage: a model with good coverage places at least one sample close to the ground truth, and needs fewer samples to do so. We report $K \in \{1, 5, 20\}$, which quantifies how quickly additional samples improve coverage.

\subsection{Density Calibration}

\begin{figure}[hbt!]
\centering
\includegraphics[width=1\textwidth]{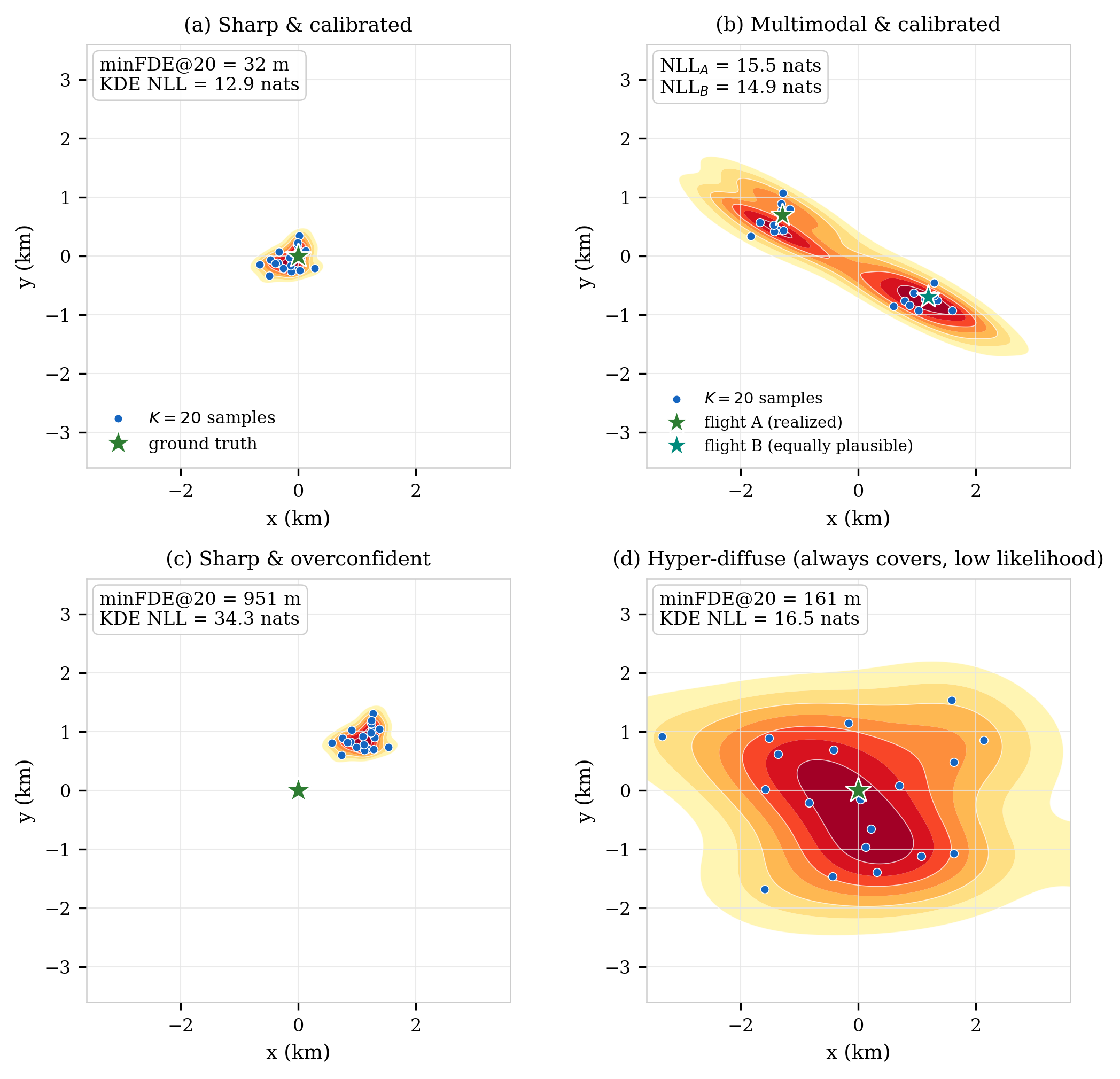}
\caption{Four synthetic
  $K{=}20$ sample-cloud scenarios, with the actual minFDE@20 and KDE NLL computed via Eq.~\eqref{eq:kde}. (a)~Sharp and calibrated: both metrics agree. (b)~Multimodal and calibrated: two equally plausible maneuvers; the bimodal density scores well against \emph{either} outcome (NLL$_A{=}15.5$, NLL$_B{=}14.9$ nats), (c)~Sharp but displaced: over-confidence makes the NLL diverge even though the samples are tightly clustered. (d)~Hyper-diffuse: spreading the samples widely keeps minFDE@20 comparable to (b), yet the likelihood assigned to the true outcome is markedly lower everywhere.}

\label{fig:kde_ade_illustration}
\end{figure}

Best-of-$K$ displacement errors reward coverage, whether at least one sample lands near the ground truth, but say nothing about how the remaining probability mass is distributed \cite{thiede2019variety}. A model that places one sample on target and scatters the other $K-1$ arbitrarily attains the same $\text{minADE@}K$ as one whose entire sample cloud tightly brackets the truth, yet only the latter yields a density usable for conflict detection (Fig.~\ref{fig:kde_ade_illustration}). To assess the quality of the full predicted distribution, we adopt the kernel-density negative log-likelihood (KDE-NLL), a widely used distributional evaluation metric in multimodal trajectory forecasting~\cite{salzmann2020trajectron}.

At a future step $n$ we draw $K=50$ independent completions and retain their horizontal positions $\{\hat{\mathbf{p}}^{(k)}_n\}_{k=1}^{K}\subset\mathbb{R}^2$. We fit a Gaussian kernel density estimate with Scott's-rule bandwidth,
\begin{equation}
  \hat{f}_n(\mathbf{p}) =
  \frac{1}{K}\sum_{k=1}^{K}
  \mathcal{N}\!\bigl(\mathbf{p};\,\hat{\mathbf{p}}^{(k)}_n,\,
  h_n^2\,\hat{\boldsymbol{\Sigma}}_n\bigr),
  \qquad h_n = K^{-1/(d+4)} = K^{-1/6},
  \label{eq:kde}
\end{equation}

where $d=2$, $\hat{\boldsymbol{\Sigma}}_n$ is the empirical covariance of the $K$ samples, and $h_n$ is the Scott factor.\footnote{A minimum bandwidth floor of 50\,m per dimension is applied so that a degenerate (collapsed) sample cloud does not drive the NLL to $-\infty$.} We then report the negative
log-likelihood of the ground-truth position under this density,
\begin{equation}
  \text{NLL@}n = -\log \hat{f}_n(\mathbf{p}_n),
  \label{eq:nll}
\end{equation}
averaged over the test set (lower is better).

The log score is strictly proper~\cite{gneiting2007scoring}. The finite-$K$ kernel estimate used here approximates it, with a Scott bandwidth and a floor, so we treat it as an approximate log-density score that jointly reflects sharpness and calibration rather than as an exactly proper rule. It rewards sharpness (concentrating mass), but penalizes over-confidence, since a tight cluster that excludes the ground truth drives $\hat{f}_n(\mathbf{p}_n)\to 0$ and the score diverges. It thus captures exactly what best-of-$K$ misses: whether the model assigns calibrated probability to where the aircraft actually goes. We evaluate NLL at the 10th, 20th, and 43rd future steps to track calibration as uncertainty accumulates over the horizon; as with the displacement metrics, the vertical axis is handled separately and the density is estimated in the horizontal plane.

\section{Baseline Architectures}
\label{sec:baselines}

We compare against three baselines spanning the kinematic, deterministic-neural, and probabilistic-generative families. All baselines consume the same 43-point observed prefix and predict the same 43-point future suffix, enabling a like-for-like comparison.

\subsection{Constant Velocity}

The \textit{constant-velocity} (CV) model propagates the last observed state forward at constant velocity. Let $t_n$ denote the (irregular) timestamp of the sample at sequence index $n$, and let $T_{\text{obs}}$ index the last observed sample. The predicted position at sequence index $T_{\text{obs}}+k$ is
\begin{equation}
  \hat{\mathbf{p}}_{T_{\text{obs}}+k}
  = \mathbf{p}_{T_{\text{obs}}}
  + \bigl(t_{T_{\text{obs}}+k} - t_{T_{\text{obs}}}\bigr)\,
    \mathbf{v}_{T_{\text{obs}}},
  \qquad k = 1,\dots,T_{\text{fut}},
  \label{eq:cv}
\end{equation}
where $\mathbf{p}_{T_{\text{obs}}}$ and $\mathbf{v}_{T_{\text{obs}}}$ are the last observed position and velocity. The elapsed time $t_{T_{\text{obs}}+k}-t_{T_{\text{obs}}}$ is read from the target timestamps and therefore accounts for the irregular ADS-B sampling, while the position index $T_{\text{obs}}+k$ remains a discrete sequence index. CV requires no training, is interpretable, and is the natural lower bound that any structure-aware model must beat. Being deterministic, its best-of-$K$ metrics are constant in $K$.

\subsection{Deterministic LSTM Encoder--GRU Decoder}

The deterministic neural baseline is a recurrent encoder--decoder. An LSTM encoder ingests the 43 observed tokens and produces a context vector; a GRU decoder then autoregressively rolls out the 43 future states. At each decoder step, the elapsed time since the start of the observation window is concatenated to the decoder input, letting the network condition its rollout on the irregular ADS-B sampling rather than assuming a fixed step. The model is trained with mean-squared error on the future per-step displacements (deltas), integrated at inference to recover absolute Cartesian positions. Because the output is a single trajectory, the model is deterministic and its @$K$ metrics again collapse to the $K=1$ values.

\subsection{Conditional Variational Autoencoder (Trajectron++)}

The strongest non-diffusion baseline is a Conditional Variational Autoencoder in the style of Trajectron++~\cite{salzmann2020trajectron}, a widely used probabilistic approach to multimodal trajectory forecasting. A recurrent encoder summarizes the observed past into a conditioning vector; a latent variable captures the discrete and continuous modes of the future (e.g.\ turn vs.\ straight); and a recurrent decoder generates a future trajectory conditioned jointly on the past and a latent sample. Drawing $K$ independent latent samples yields $K$ diverse trajectories, so, unlike CV and the deterministic LSTM, the CVAE supports genuine best-of-$K$ evaluation and density estimation, and serves as our probabilistic non-diffusion reference.

\section{FlowATC: Generative Inpainting with Flow Matching}
\label{sec:approach}

\subsection{Problem Formulation}
\label{sec:formulation}

Let $(\xobs, \xfut) \sim p_{\text{data}}$ denote a pair of observed prefix and future suffix drawn from the (unknown) joint distribution of Bay Area traffic, with $\xobs \in \Rbb^{T_{\text{obs}}\times F}$ and $\xfut \in \Rbb^{T_{\text{fut}}\times F}$. Because runway assignment, controller instructions, and traffic sequencing are not observable in $\xobs$, the conditional law $p(\xfut \mid \xobs)$ is in general multimodal: several distinct futures carry non-negligible probability mass. The object we seek is therefore not a point estimate but a sampler for this conditional distribution, that is, a mechanism producing $\hat{\mathbf{X}}^{\text{fut}} \sim p(\cdot \mid \xobs)$, from which any downstream quantity (conflict probability, occupancy density, best of $K$ forecasts) can be estimated by Monte Carlo.

This requirement is not merely a preference: any deterministic predictor $f_\theta$ trained with mean squared error converges, at the population optimum, to the conditional mean,
\begin{equation}
  \arg\min_{f}\;
  \mathbb{E}_{(\xobs,\xfut)}
  \bigl\|f(\xobs) - \xfut\bigr\|_2^2
  \;=\;
  \mathbb{E}\bigl[\xfut \mid \xobs\bigr],
  \label{eq:cond_mean}
\end{equation}
by the standard $L^2$ projection property of conditional expectation. When $p(\xfut\mid\xobs)$ has two modes, say a left turn and a continued straight leg, their average is a trajectory that belongs to neither mode and may be physically implausible. The deterministic baselines of Section~\ref{sec:baselines} are thus limited by construction, independently of their capacity: they solve a different (and, under multimodality, ill-suited) problem.

Conditional Flow Matching sidesteps this by regressing a velocity field $u_\theta$ rather than a trajectory. Fix $\xobs$, draw $\mathbf{x}_0 \sim \mathcal{N}(\mathbf{0},\mathbf{I})$ and $\mathbf{x}_1 = \xfut \sim p(\cdot\mid\xobs)$, and define the linear interpolant $\mathbf{x}_t = (1-t)\,\mathbf{x}_0 + t\,\mathbf{x}_1$. The population CFM objective regresses $u_\theta(\mathbf{x}_t, t \mid \xobs)$ onto the pair conditional velocity $(\mathbf{x}_1 - \mathbf{x}_0)$. Because the squared loss is minimized pointwise by a conditional expectation, its unique population minimizer is the marginal velocity field
\begin{equation}
  u^\star(\mathbf{x}, t \mid \xobs)
  \;=\;
  \mathbb{E}\bigl[\,\mathbf{x}_1 - \mathbf{x}_0
    \;\big|\; \mathbf{x}_t = \mathbf{x},\; \xobs \bigr],
  \label{eq:marginal_field}
\end{equation}
and it is a standard result of the flow matching literature~\cite{lipman2022flow,albergo2023stochastic} that the probability flow of this field, namely the solution of $\dot{\mathbf{x}} = u^\star(\mathbf{x},t\mid\xobs)$ initialized at $\mathbf{x}(0)\sim\mathcal{N}(\mathbf{0},\mathbf{I})$, has marginal law exactly $p(\cdot\mid\xobs)$ at $t=1$. In other words, exactly minimizing the CFM loss and exactly integrating the learned field is equivalent to sampling from the true conditional distribution of futures. Multimodality is preserved automatically: distinct noise draws $\mathbf{x}_0$ are transported to distinct modes, and no averaging across modes ever occurs.

\subsection{The DiT Backbone}

Following the Diffusion Transformer (DiT) framework~\cite{peebles2023dit}, the denoiser is a stack of Transformer blocks with the generation step (flow time $t\in[0,1]$) injected into every block through Adaptive Layer Normalization (AdaLN). Let $\mathbf{h}\in\Rbb^d$ denote a token's hidden representation entering a sub-layer, with $\mu(\mathbf{h})$ and $\sigma(\mathbf{h})$ its mean and standard deviation taken over the feature dimension $d$ (per-token normalization). A small MLP maps $t$ to scale and shift parameters $\gamma(t),\beta(t)\in\Rbb^d$, applied via the same AdaLN mechanism at each of the two sub-layers (self-attention and feed-forward) through independently learned MLP heads:
\begin{equation}
  \text{AdaLN}(\mathbf{h}, t) =
    \gamma(t)\cdot
    \frac{\mathbf{h} - \mu(\mathbf{h})}{\sigma(\mathbf{h})}
    + \beta(t).
\end{equation}
The core block operation is multi-head self-attention; for tokens $\mathbf{Z}\in\Rbb^{T\times d}$, split into $H$ heads of dimension $d_k = d/H$,
\begin{equation}
  \text{Attn}(\mathbf{Z}) =
  \text{softmax}\!\left(\frac{\mathbf{Q}\mathbf{K}^\top}{\sqrt{d_k}}\right)\mathbf{V},
  \quad
  \mathbf{Q},\mathbf{K},\mathbf{V} = \mathbf{Z}W_Q,\,\mathbf{Z}W_K,\,\mathbf{Z}W_V,
\end{equation}

Attention is block-causal: observed tokens are prevented from attending to the noisy future tokens, while future tokens attend freely to the observed prefix and to one another. Writing $\mathbf{A}\in\{0,-\infty\}^{T\times T}$ for the additive mask applied to the attention logits before the softmax, $A_{ij}=-\infty$ iff $i<T_{\text{obs}}$ and $j\geq T_{\text{obs}}$, and $0$ otherwise. The observed representation is therefore independent of the noise realisation, while the future block keeps the full bidirectional attention motivated above, since all its tokens share the same noise level. The future timestamps supplied through $t_{\text{rel}}$ are query times: they state when a prediction is requested and carry no information about the aircraft's future state.

\begin{figure}[hbt!]
    \centering
    \includegraphics[width=0.9\linewidth]{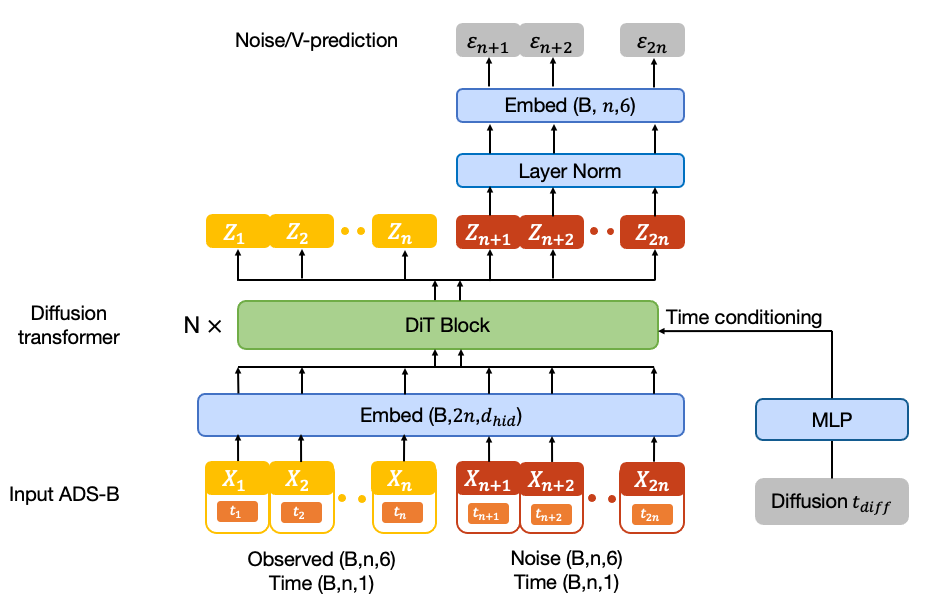}
    \caption{FlowATC DiT\cite{peebles2023dit} architecture overview.}
    \label{fig:architecture}
\end{figure}

Figure~\ref{fig:architecture} shows the full architecture.

\subsection{Trajectory as an Image: Inpainting the Future}

\begin{figure}
    \centering
    \includegraphics[width=0.75\linewidth]{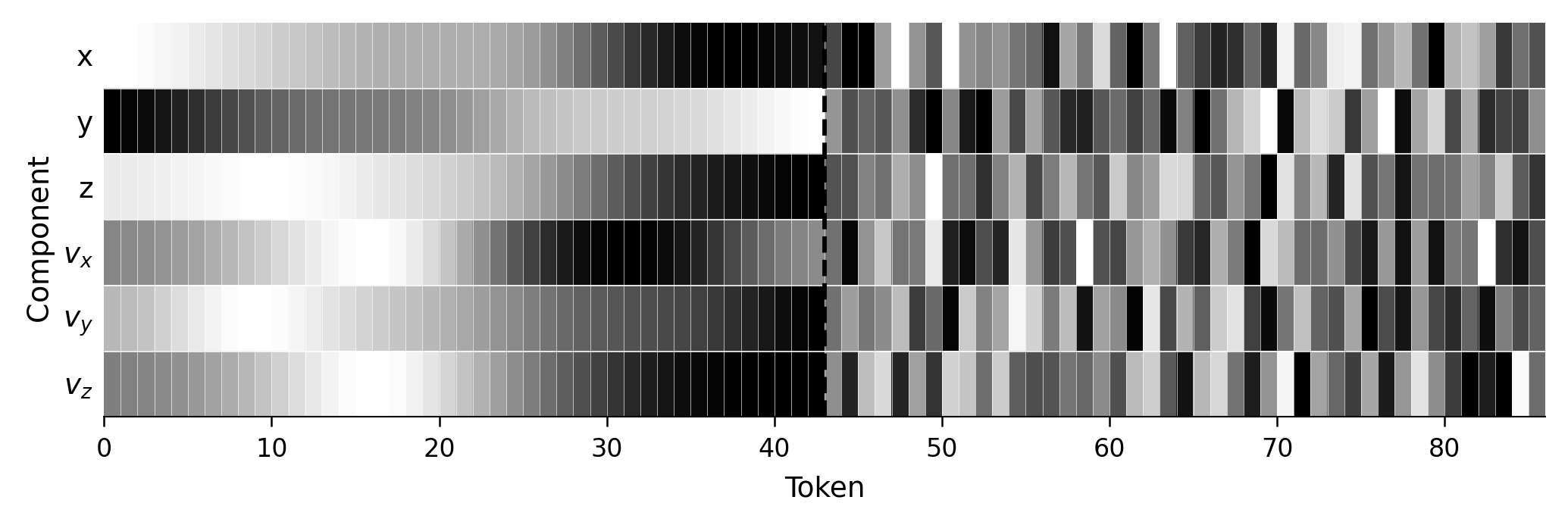}
    \caption{Trajectory represented as a $6\times 86$ feature-time matrix. The observed prefix (solid, left of dashed line) provides context tokens; the noisy future suffix (light, right) is the region to be inpainted by the flow-matching model.}
    \label{fig:inpaint_concept}
\end{figure}

We represent each 86-step window as a $6\times86$ feature-time matrix, in effect a one-channel ``image'' of the trajectory. The first $T_{\text{obs}}=43$ tokens carry the observed (clean) state vectors; the last $T_{\text{fut}}=43$ tokens are initialized with Gaussian noise and treated as the masked region to be inpainted (Fig.~\ref{fig:inpaint_concept}).

Both the $T_{\text{obs}}$ clean observation tokens and the $T_{\text{fut}}$ noisy future tokens are embedded by a shared linear projection and concatenated into a single sequence of length $T=86$:
\begin{equation}
  \mathbf{Z} = \bigl[W_{\text{in}}\,\xobs \;\|\;
                     W_{\text{in}}\,\tilde{\mathbf{X}}^{\text{fut}}\bigr]
  \in \Rbb^{T \times d}.
  \label{eq:concat}
\end{equation}
The observed past is thus injected purely by concatenation: no separate encoder is needed. Because the past enters only as additional tokens, the architecture is a natural substrate for further conditioning: any extra signal (ATC voice, weather fields, charts) can be appended as context tokens without changing the loss or the inpainting mechanism. Because all future tokens are denoised at the same noise level $t$, there is no temporal ordering to preserve within the future block, so attending freely helps the model produce spatially consistent trajectories. Only the $T_{\text{fut}}$ future positions are read off for the loss or the next integration step.

In addition to the AdaLN diffusion-time conditioning, each token receives a sinusoidal time embedding of its prediction date: the time offset of that token within the window which is added to the token representation. This lets the model distinguish near-future from far future positions even though all future tokens are denoised simultaneously.

The velocity field $u_\theta(\mathbf{x},t)$ is trained on the straight-line interpolation between Gaussian noise and the target future:
\begin{equation}
  \mathcal{L}_{\text{CFM}} =
  \mathbb{E}_{t,\,\mathbf{x}_0,\,\xfut}
  \Bigl\|
    u_\theta\!\bigl((1-t)\mathbf{x}_0 + t\xfut,\; t \mid \xobs\bigr)
    - (\xfut - \mathbf{x}_0)
  \Bigr\|_2^2,
  \label{eq:cfm}
\end{equation}
with $\mathbf{x}_0\sim\mathcal{N}(\mathbf{0},\mathbf{I})$ and
$t\sim\text{LogitNormal}(1.0,1.0)$.
At inference, Euler integration from $t=0$ to $t=1$ in 20 steps produces one trajectory sample; repeating with independent noise draws yields $K$ diverse completions. We refer to this flow-matching-trained model as FlowATC. The same backbone is also trained with a DDPM objective~\cite{ho2020ddpm} (1000 forward steps, cosine schedule, DDIM sampling) as a baseline.

All CFM and DDPM variants share a Transformer backbone (multi-head self-attention, feed-forward factor 4, dropout 0.1) with the layer/head/width counts of Table~\ref{tab:arch_hp}. Training uses AdamW ($\beta_1=0.9$, $\beta_2=0.999$, weight decay $10^{-2}$, learning rate $10^{-4}$) with a 1000-step linear warm-up, batch size 512, for up to 200 epochs on a single GPU. Exponential Moving Average weights (decay 0.9999) are used at inference.

Latency is the wall-clock time to answer one request: a single input sequence (batch size 1) for which $K{=}20$ samples are drawn in parallel, using 20 Euler steps for flow models and 100 DDIM steps for diffusion models. It is measured in float32 on a single NVIDIA RTX 4000 Ada, averaged over at least 30 requests after
10 warm-up requests.

\begin{table}[hbt!]
\caption{Flow model architecture configurations.}
\label{tab:arch_hp}
\centering
\small
\begin{tabular}{lcccc}
\toprule
Model & Layers & Heads & $d$ & Parameters \\
\midrule
flow\_tiny & 5  & 4 & 128 & 1.5\,M \\
flow\_small & 6  & 8 & 256 & 7.1\,M \\
flow\_large & 8 & 8 & 384 & 20.7\,M \\
\bottomrule
\end{tabular}
\end{table}

\section{Results}
\label{sec:results}

\begin{figure}[t!]
\centering
\includegraphics[width = 0.65\linewidth]{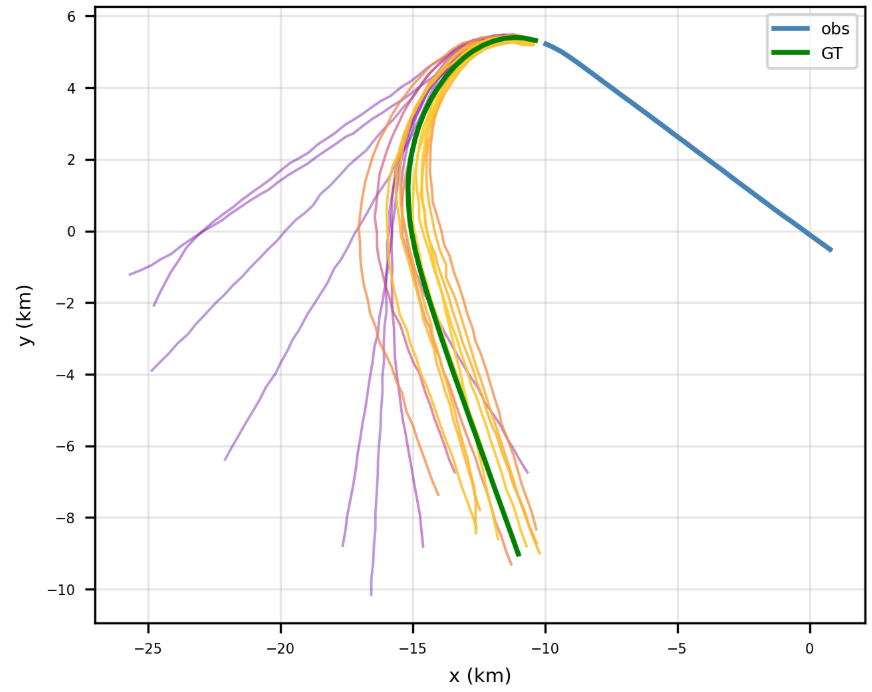}
\caption{Predicted trajectory fan ($K=20$ samples). Blue: observed past; Green: ground-truth future; purple--yellow: predicted final-point density (purple = low, yellow = high).}
\label{fig:fan_plot}
\end{figure}

\begin{figure}[hbt!]
\centering
\begin{subfigure}[t]{0.32\textwidth}
  \centering
  \includegraphics[width=\linewidth]{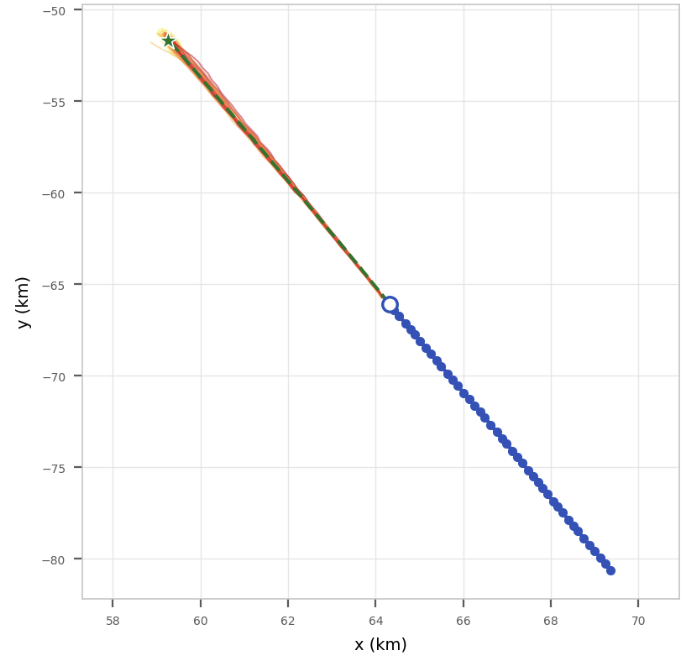}
  \caption{Straight en-route}
  \label{fig:acc_enroute}
\end{subfigure}
\hfill
\begin{subfigure}[t]{0.315\textwidth}
  \centering
  \includegraphics[width=\linewidth]{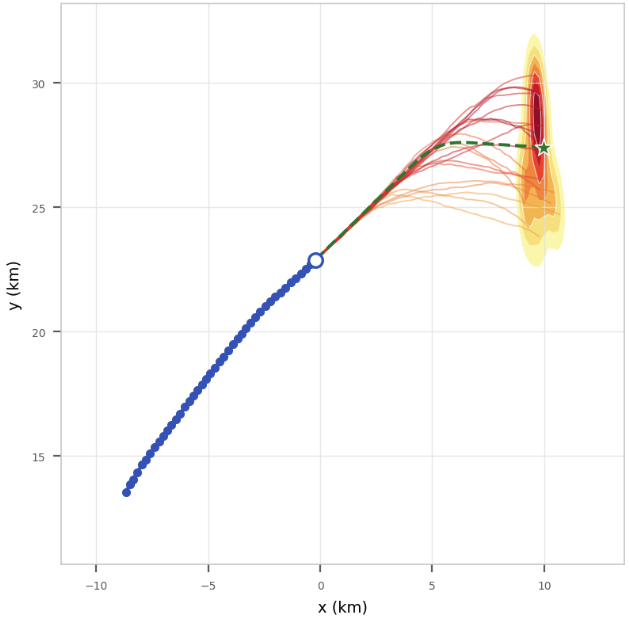}
  \caption{Turn}
  \label{fig:acc_turn}
\end{subfigure}
\hfill
\begin{subfigure}[t]{0.325\textwidth}
  \centering
  \includegraphics[width=\linewidth]{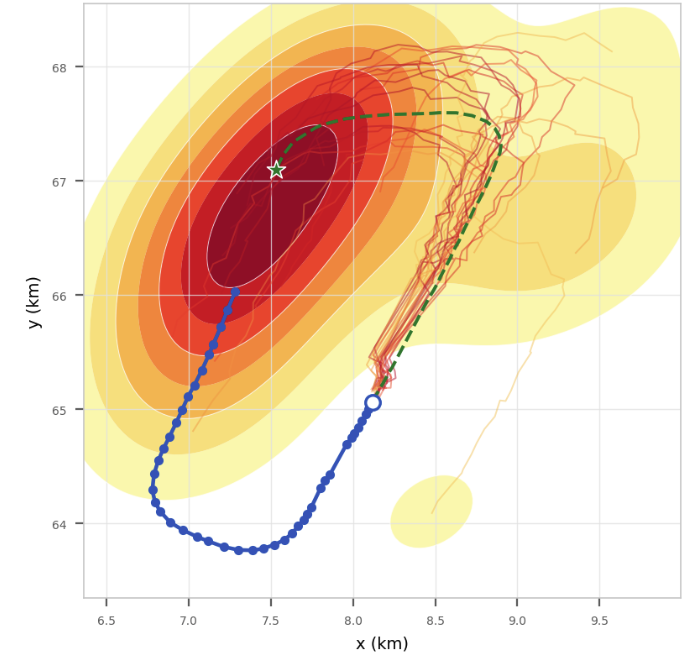}
  \caption{Holding entry}
  \label{fig:acc_hold}
\end{subfigure}

\caption{Final-position probability density maps ($K=20$ samples, Gaussian KDE, Scott bandwidth). Dashed blue: observed past; dashed green: ground-truth future; solid heatmap: sampled trajectories; green star: ground-truth final position; yellow--red: predicted final-point density (light = low, dark = high). Density width correlates with the complexity of the anticipated maneuver.}
\label{fig:heatmap}
\end{figure}

Drawing $K$ completions and coloring each by its final-point KDE density yields a fan of plausible futures (Fig.~\ref{fig:fan_plot}); the corresponding final-point density forms spatial probabilistic occupancy estimates that can serve as an input to downstream conflict-risk estimation (Fig.~\ref{fig:heatmap}). For straight en-route segments the density is narrow and elongated along the track; for aircraft entering an approach turn it broadens laterally, capturing genuine uncertainty about the turn-initiation point. In all cases the high-density region brackets the ground-truth final position.

\subsection{Model Comparison}
\begin{figure}[t!]
\centering
\includegraphics[width = 1.0\linewidth]{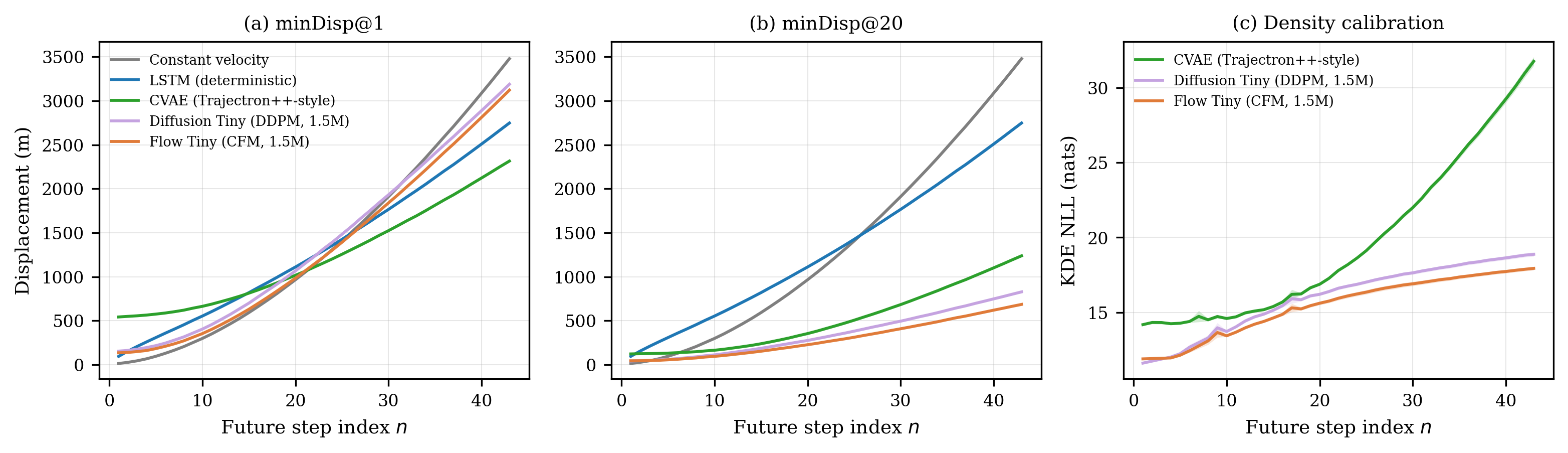}
\caption{Prediction error and calibration across sample budget and future horizon. (a)–(b) Minimum displacement error at $K\in\{1,20\}$. (c) Density calibration for the two stochastic generative models (CV and the deterministic LSTM are excluded: their point estimate has no associated density). The full $K\in{1,5,20}$ sweep is given in Appendix~\ref{app:degradation_full}, Fig.~\ref{fig:degradation_full}.}

\label{fig:degradation}
\end{figure}

Table~\ref{tab:model_comparison} compares all five model variants at matched capacity (1.5\,M parameters) on the held-out test set. LSTM-Det improves over CV mainly on FDE (2748.5\,m vs.\ 3480.2\,m, $-21\%$), reflecting that a learned decoder curves its rollout toward typical approach geometry rather than extrapolating a straight line; the ADE gain is comparatively modest (1305.3\,m vs.\ 1343.6\,m), since most of the 43-step horizon is still well approximated by locally linear motion. At $K=1$, the generative models do \emph{not} outperform the deterministic baselines outright: Diffusion Tiny (1370.6/3183.3) is in fact the worst model in the table on minADE@1, and Flow Tiny (1313.5/3147.7) is comparable to CV. Panel (a) of Fig.~\ref{fig:degradation} shows this is a genuine crossover in horizon, not just an artefact of the final step: CV, LSTM, Diffusion, and Flow all start near zero displacement and grow with $n$, while CVAE starts at $\sim$500\,m even at $n{=}1$ and grows far more slowly, overtaking the other four models only around $n\approx15$–20. This is expected rather than a failure mode: a single diffusion or flow draw is one stochastic sample from the predicted distribution, not a mean or mode estimate, so it inherits the full spread of the multimodal posterior from the very first future step. CVAE's early offset has a different origin, isolated in Appendix~\ref{app:cvae}: it is decoder output variance rather than a mispredicted mean, and the model cannot shed this noise for free as doing so collapses minADE@20 instead, since its discrete latent supplies little diversity conditional on a single observed history. The generative models only become competitive once they are allowed to hedge across samples.

\begin{table}[hbt!]
\caption{Model comparison at equal capacity (1.5\,M parameters). 43\,obs $\to$ 43\,fut steps ($\approx128$\,s horizon), $N=63{,}016$ test trajectories. Distances in metres~$\pm$\,SEM\@. KDE NLL in nats (lower is better for all columns). \textbf{Bold} = best value per column.}

\label{tab:model_comparison}
\centering
\footnotesize
\setlength{\tabcolsep}{4pt}
\resizebox{\textwidth}{!}{%
\begin{tabular}{l l c c cc cc cc ccc}
\toprule
Model & Type & Params & ms/pred
  & \multicolumn{2}{c}{$K=1$}
  & \multicolumn{2}{c}{$K=5$}
  & \multicolumn{2}{c}{$K=20$}
  & \multicolumn{3}{c}{KDE NLL@$n$}\\
\cmidrule(lr){5-6}\cmidrule(lr){7-8}\cmidrule(lr){9-10}\cmidrule(lr){11-13}
 & & & & minADE & minFDE & minADE & minFDE & minADE & minFDE
   & $n{=}10$ & $n{=}20$ & $n{=}43$ \\
\midrule
CV                           & ---  & ---  &  $0.0$     & $1343.6_{\pm8.1}$          & $3480.2_{\pm19.3}$          & $1343.6_{\pm8.1}$          & $3480.2_{\pm19.3}$          & $1343.6_{\pm8.1}$          & $3480.2_{\pm19.3}$          & ---                       & ---                       & --- \\
LSTM-Det ($+t_{\rm rel}$)  & ---  & 1.5\,M & $1.15$ & $1305.3_{\pm5.7}$          & $2748.5_{\pm12.6}$          & $1305.3_{\pm5.7}$          & $2748.5_{\pm12.6}$          & $1305.3_{\pm5.7}$          & $2748.5_{\pm12.6}$          & ---                       & ---                       & ---\\
\addlinespace
CVAE (Trajectron++ style)     & CVAE & 1.5\,M &  $4.06$ & $\mathbf{1219.7}_{\pm6.0}$ & $\mathbf{2316.2}_{\pm12.8}$ & $828.5_{\pm4.3}$           & $1608.8_{\pm10.2}$          & $662.9_{\pm3.8}$           & $1240.9_{\pm9.1}$           & $14.6_{\pm0.0}$           & $16.9_{\pm0.1}$           & $31.7_{\pm0.3}$ \\
Diffusion Tiny                  & DDPM & 1.5\,M &  $119.7$ & $1370.6_{\pm7.5}$          & $3183.3_{\pm17.6}$          & $697.1_{\pm4.2}$           & $1467.5_{\pm9.3}$           & $459.9_{\pm2.9}$           & $828.1_{\pm6.1}$            & $13.7_{\pm0.0}$           & $16.2_{\pm0.1}$           & $18.9_{\pm0.1}$ \\
Flow Tiny                       & CFM  & 1.5\,M &  $23.8$ & $1313.5_{\pm7.5}$          & $3147.7_{\pm18.2}$          & $\mathbf{614.5}_{\pm3.8}$  & $\mathbf{1302.7}_{\pm8.7}$  & $\mathbf{390.5}_{\pm2.4}$  & $\mathbf{686.1}_{\pm5.0}$   & $\mathbf{13.4}_{\pm0.1}$  & $\mathbf{15.7}_{\pm0.1}$  & $\mathbf{18.1}_{\pm0.2}$ \\
\bottomrule
\end{tabular}}
\end{table}

The picture reverses sharply by $K=20$: Flow Tiny (390.5/686.1) and Diffusion Tiny (459.9/828.1) both surpass CVAE (662.9/1240.9) by a wide margin, $41\%$ and $31\%$ lower minADE@20 respectively. Panel (b) of Fig. \ref{fig:degradation} shows this advantage holds across the full horizon, not just at $n{=}43$: flow and diffusion separate from CVAE as early as $K{=}5$ (Appendix~\ref{app:degradation_full}), and the gap widens monotonically with $n$ through $K{=}20$. This indicates that the diffusion/flow sample cloud covers the true multimodal distribution more efficiently than the CVAE's discrete-latent mixture; fewer samples are wasted on implausible modes, so additional draws pay off faster at every horizon, not only at the end of the window.

The NLL@$n$ columns tell a complementary story about distributional quality rather than best-case coverage. All three generative models are similar at the 10-step horizon (13.4--14.6\,nats), but diverge sharply by the 43-step horizon, as panel (c) of Fig.~\ref{fig:degradation} makes clear: CVAE's calibration curve bends upward steeply after $n\approx20$, while Diffusion Tiny and Flow Tiny remain nearly flat over the same range. CVAE's NLL@43 (31.7) corresponds to a kernel density at the realized endpoint roughly $e^{13.6}\approx8\times10^{5}$ times smaller than Flow Tiny's (18.1), meaning that even though CVAE's best-of-20 samples can land close to the ground truth, its \emph{full} predicted density is comparatively poorly calibrated at long horizons.

\begin{table}[hbt!]
\centering
\caption{Vertical error at matched capacity (1.5\,M parameters) on the held-out test set, in metres, mean $\pm$ SEM over 63,016 windows. The altitude axis is evaluated separately from the horizontal metrics of Table~\ref{tab:model_comparison}; the deterministic baselines are constant in $K$.}
\label{tab:vertical}
\begin{tabular}{l cc cc}
\toprule
Model
  & \multicolumn{2}{c}{minADE$_z$}
  & \multicolumn{2}{c}{minFDE$_z$} \\
\cmidrule(lr){2-3}\cmidrule(lr){4-5}
 & $K{=}1$ & $K{=}20$ & $K{=}1$ & $K{=}20$ \\
\midrule
CV              & $109.7_{\pm0.7}$         & $109.7_{\pm0.7}$         & $248.7_{\pm1.5}$          & $248.7_{\pm1.5}$ \\
LSTM-Det        & $\mathbf{78.4}_{\pm0.4}$ & $78.4_{\pm0.4}$          & $\mathbf{151.9}_{\pm0.9}$ & $151.9_{\pm0.9}$ \\
CVAE            & $94.6_{\pm0.6}$          & $43.8_{\pm0.3}$          & $168.6_{\pm1.2}$          & $62.0_{\pm0.7}$ \\
Diffusion Tiny  & $102.6_{\pm0.5}$         & $28.3_{\pm0.2}$          & $198.9_{\pm1.1}$          & $29.9_{\pm0.4}$ \\
Flow Tiny       & $94.2_{\pm0.5}$          & $\mathbf{25.0}_{\pm0.2}$ & $186.8_{\pm1.1}$          & $\mathbf{24.3}_{\pm0.3}$ \\
\bottomrule
\end{tabular}
\end{table}

Comparing the two generative objectives directly, CFM (Flow Tiny) beats DDPM (Diffusion Tiny) on every column: $15\%$ lower minADE@20 (390.5 vs.\ 459.9), $17\%$ lower minFDE@20, and consistently lower NLL at all three horizons. In this matched-capacity comparison CFM consistently outperforms DDPM; since capacity is held fixed, the gap points to the training objective rather than to model size, though we report single runs and do not quantify seed variance, a point we revisit at scale in Section~\ref{sec:results-scaling}.

Table~\ref{tab:vertical} reports the vertical axis, evaluated separately. It reproduces the horizontal narrative rather than adding a new one. At $K{=}1$ the deterministic LSTM is the most accurate model ($78.4_{\pm0.4}$\,m against $94.2_{\pm0.5}$\,m for Flow Tiny), since a single stochastic draw again inherits the spread of the predicted distribution rather than estimating its mean. By $K{=}20$ the ordering reverses: Flow Tiny reaches $25.0_{\pm0.2}$\,m against $28.3_{\pm0.2}$\,m for Diffusion Tiny and $43.8_{\pm0.3}$\,m for the CVAE, a $43\%$ reduction over the latter, close to the $41\%$ measured horizontally. Altitude at a two-minute horizon is dominated by the discrete choice to level off, climb or descend, precisely the kind of branching a single point estimate cannot represent.

\subsection{Scaling}
\label{sec:results-scaling}

\begin{table}[hbt!]
\caption{Flow Matching vs.\ Diffusion scaling. 43\,obs $\to$ 43\,fut steps ($\approx128$\,s horizon), $N=63{,}016$ test trajectories. Distances in metres~$\pm$\,SEM\@. KDE NLL in nats. \textbf{Bold} = best within each architecture family.}
\label{tab:scaling}
\centering
\footnotesize
\setlength{\tabcolsep}{4pt}
\resizebox{\textwidth}{!}{%
\begin{tabular}{l l c cc cc cc ccc}
\toprule
Model & Type & Params
  & \multicolumn{2}{c}{$K=1$}
  & \multicolumn{2}{c}{$K=5$}
  & \multicolumn{2}{c}{$K=20$}
  & \multicolumn{3}{c}{KDE NLL@$n$} \\
\cmidrule(lr){4-5}\cmidrule(lr){6-7}\cmidrule(lr){8-9}\cmidrule(lr){10-12}
 & & & minADE & minFDE & minADE & minFDE & minADE & minFDE
   & $n{=}10$ & $n{=}20$ & $n{=}43$ \\
\midrule
\multicolumn{12}{l}{\textit{Flow Matching (CFM, 20 inference steps)}} \\[2pt]
Flow Tiny  & CFM & 1.5\,M  & $1313.5_{\pm7.5}$         & $3147.7_{\pm18.2}$         & $614.5_{\pm3.8}$          & $1302.7_{\pm8.7}$          & $390.5_{\pm2.4}$          & $686.1_{\pm5.0}$          & $13.4_{\pm0.1}$           & $\mathbf{15.7}_{\pm0.1}$  & $\mathbf{18.1}_{\pm0.2}$ \\
Flow Small & CFM & 7.1\,M  & $996.0_{\pm6.5}$           & $2419.7_{\pm15.8}$         & $479.5_{\pm3.4}$          & $1050.6_{\pm7.9}$          & $\mathbf{306.1}_{\pm2.2}$ & $\mathbf{568.7}_{\pm4.7}$ & $\mathbf{13.3}_{\pm0.1}$  & $17.4_{\pm0.6}$           & $20.2_{\pm0.5}$ \\
Flow Large & CFM & 20.7\,M & $\mathbf{903.4}_{\pm5.9}$  & $\mathbf{2156.8}_{\pm14.2}$& $\mathbf{467.3}_{\pm3.4}$ & $\mathbf{1013.2}_{\pm7.8}$ & $307.4_{\pm2.4}$          & $576.8_{\pm5.3}$          & $13.3_{\pm0.1}$           & $17.5_{\pm0.5}$           & $20.2_{\pm0.4}$ \\
\addlinespace
\multicolumn{12}{l}{\textit{Diffusion (DDPM, 100 inference steps)}} \\[2pt]
Diffusion Tiny  & DDPM & 1.5\,M  & $1370.6_{\pm7.5}$          & $3183.3_{\pm17.6}$          & $697.1_{\pm4.2}$           & $1467.5_{\pm9.3}$           & $459.9_{\pm2.9}$           & $828.1_{\pm6.1}$           & $13.7_{\pm0.0}$           & $\mathbf{16.2}_{\pm0.1}$  & $\mathbf{18.9}_{\pm0.1}$ \\
Diffusion Small & DDPM & 7.1\,M  & $1173.8_{\pm6.9}$          & $2835.5_{\pm16.5}$          & $620.2_{\pm4.0}$           & $1385.8_{\pm9.3}$           & $412.7_{\pm2.8}$           & $805.9_{\pm6.3}$           & $14.2_{\pm0.1}$           & $19.1_{\pm0.3}$           & $22.6_{\pm0.4}$ \\
Diffusion Large & DDPM & 20.7\,M & $\mathbf{1051.6}_{\pm6.3}$ & $\mathbf{2478.6}_{\pm15.3}$ & $\mathbf{529.8}_{\pm3.5}$  & $\mathbf{1150.3}_{\pm8.2}$  & $\mathbf{344.0}_{\pm2.4}$  & $\mathbf{650.3}_{\pm5.5}$  & $\mathbf{13.2}_{\pm0.1}$  & $16.8_{\pm0.3}$           & $20.9_{\pm0.6}$ \\
\bottomrule
\end{tabular}}
\end{table}

Table~\ref{tab:scaling} reports both CFM and DDPM across three model sizes. FlowATC's minADE@20 improves sharply from Tiny to Small ($390.5\to306.1$\,m, $-21.6\%$) but plateaus from Small to Large ($306.1\to307.4$\,m, $+0.4\%$), and minFDE@20 shows the same pattern ($568.7\to576.8$\,m). This suggests CFM saturates the information available in the 12-day dataset by around 7\,M parameters, approaching a data-limited rather than capacity-limited regime. The saturation has an operational corollary: Flow Small reaches the same minADE@20 as Flow Large ($306.1$ vs $307.4$\,m) at $2.3\times$ lower latency ($67.0$ vs $157.1$\,ms per request, Table~\ref{tab:throughput}), so capacity beyond $7.1$\,M buys nothing but cost on this dataset.

DDPM shows the opposite trend: minADE@20 improves steadily at every step ($459.9\to412.7\to344.0$\,m, $-10.3\%$ then $-16.6\%$), with its largest single gain occurring exactly where flow stalls. Because DDPM has a harder denoising objective, this is consistent with DDPM still being capacity-limited at 20.7\,M parameters where CFM is not. Consequently, the relative advantage of CFM over DDPM on minADE@20 is not monotonic in model size: $15\%$ at Tiny, widening to $26\%$ at Small, then narrowing to $11\%$ at Large as DDPM catches up (each measured as the reduction in minADE@20 relative to DDPM at matched capacity). At the sizes tested here, CFM remains strictly better at every scale, but DDPM's steeper scaling curve suggests the gap would continue to narrow, or close, at larger capacity than we evaluate.

The KDE NLL columns decouple from minADE/minFDE in an interesting way: FlowATC's NLL@43 is \emph{best} at Tiny (18.1\,nats) and degrades with scale (20.2\,nats at both Small and Large), even as displacement error improves. Larger FlowATC models thus produce sample clouds that are more accurate on average but slightly less well calibrated at long horizons, a reminder that best-of-$K$ accuracy and distributional calibration are related but distinct properties, and that scaling helps one without guaranteeing the other.

\subsection{Extrapolation to Future Timesteps}
\label{sec:extrapolation}

Because the prediction-date embedding conditions each future token on its time offset within the window rather than on a fixed step count, the model can be queried at prediction dates beyond the $T_{\text{fut}}=43$ horizon seen during training, simply by resampling the embedding at longer offsets without retraining or architectural changes. We probe this by overwriting the $t_{\text{rel}}$ timing channel with virtual timestamps spanning target horizons up to $360$\,s (roughly three times the native $\approx\!128$\,s horizon), stitching three shards ($H_{\max}\in\{90,180,360\}$\,s) to keep good temporal resolution across the full range. Ground truth beyond the observation window is obtained by cubic-spline interpolation of the continuing raw ADS-B trajectory: a window contributes at a given horizon only if its flight actually extends that far, and horizons where fewer than half the windows qualify are dropped, a filter that does not bind here since $85\%$ of windows still have a continuing track at $360$\,s. Appendix~\ref{app:extrapolation_full} gives the full protocol, including a visible estimator-seam artefact at the shard boundaries that reflects a change of estimator, not of model behavior.

\begin{figure}[hbt!]
\centering
\includegraphics[width=0.6\linewidth]{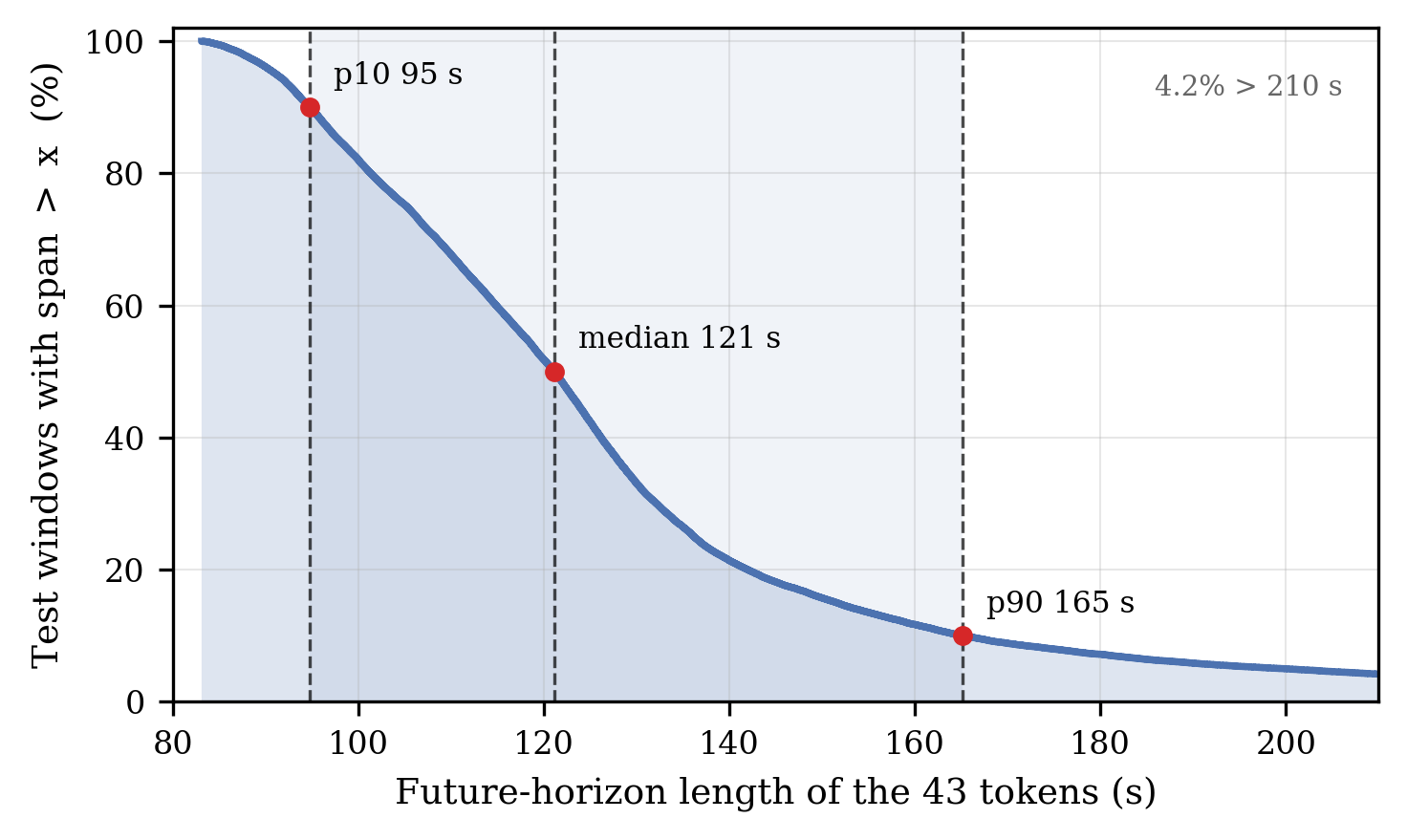}
\caption{Portion of test windows reaching a given future horizon.}
\label{fig:horizon_coverage}
\end{figure}

\begin{figure}[t!]
\centering
\includegraphics[width=\linewidth]{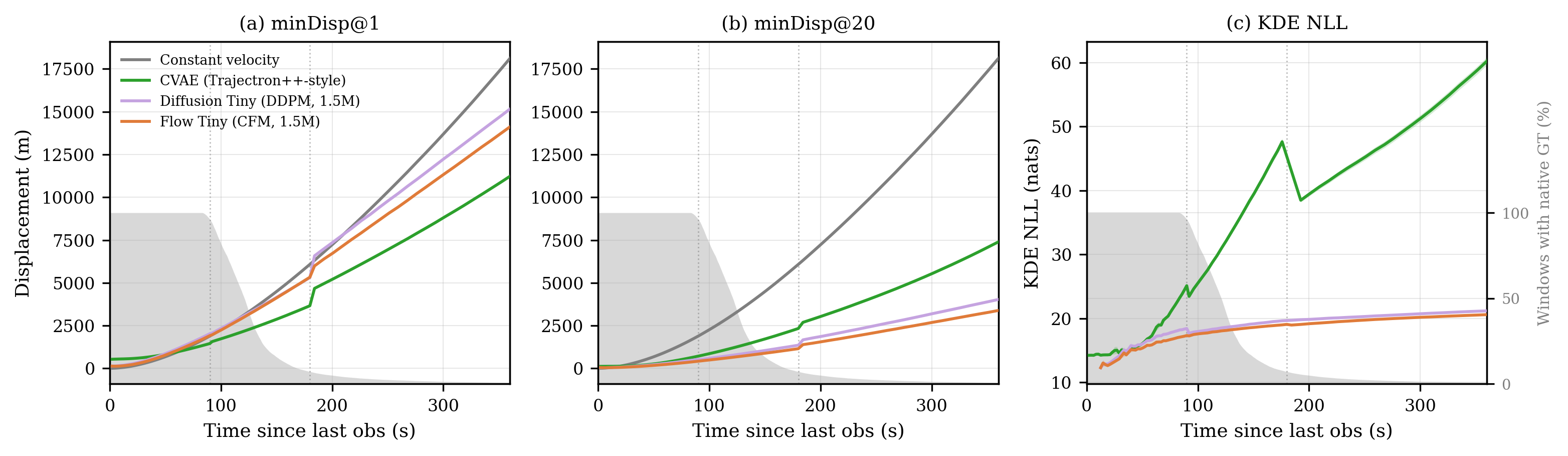}
\caption{Extrapolation error and calibration beyond the training horizon. (a)--(b) minDisp@$K$ for $K\in\{1,20\}$. (c) KDE NLL. Gray shading (right axis): fraction of test windows with native ground-truth coverage. The full $K\in\{1,5,20\}$ sweep is given in Appendix~\ref{app:extrapolation_full}.}
\label{fig:extrapolation_error}
\end{figure}

Figure~\ref{fig:extrapolation_error} reports minADE@$K$ (panels a--b) and KDE NLL (panel c) as a function of elapsed time since the last observation, from $0$ to $360$\,s, for CVAE, Diffusion Tiny, and Flow Tiny (all 1.5\,M parameters). The gray shading reports a distinct quantity: the fraction of test windows whose native 43-token future span alone reaches that horizon, $96\%$ at $90$\,s but $16\%$ at $150$\,s and under $1\%$ at $360$\,s (Fig.~\ref{fig:horizon_coverage}). Beyond roughly $150$\,s the comparison therefore rests on the interpolated continuation of each flight rather than on tokens the observation window itself contains.

The diffusion-based models degrade faster than CVAE at long horizons when only a single sample is drawn ($K{=}1$, panel a); once $K\geq5$ (panel b shows $K{=}20$, the full $K{=}5$ curve is in Appendix~\ref{app:extrapolation_full}) Flow remains the most accurate model across the full horizon, and only CVAE's $K{=}1$ curve overtakes it beyond $\approx200$\,s. This is an artefact of the models being queried at time offsets never seen during training: the future block still contains 43 tokens, but the prediction dates attached to them lie beyond the training horizon, whereas CVAE's autoregressive decoder rolls out to arbitrary length by construction.

Panel (c) shows why this apparent CVAE advantage is misleading. KDE-NLL, which tests whether a model's uncertainty keeps pace with its actual error independent of its size, plateaus at $\approx\!20$\,nats for Diffusion and Flow from $t\approx150$\,s onward. Their sample clouds widen in step with the true uncertainty even as displacement error keeps growing (panel b) while CVAE's NLL diverges to $\approx\!60$\,nats by $t=360$\,s. CVAE's lower displacement error beyond $\approx200$\,s therefore reflects a single sampled trajectory landing closer to the truth on average, not a predicted distribution that knows how uncertain it should be: exactly the failure mode best-of-$K$ metrics cannot detect and KDE-NLL is designed to catch.

\subsection{Operational Robustness}
\label{sec:robustness}

Real ADS-B receivers report at rates that vary with receiver quality and congestion. To test whether the architecture remains effective at coarser temporal resolution, we retrain the flow model at strides of 2, 4, and 8 (effective intervals $\approx5$, $10$, $20$\,s): a dedicated model per stride, not one fixed model queried at variable resolution, since obs\_len and fut\_len must change with the token count. Stride-2 costs $+4.9\%$ in minADE@20 while roughly halving inference time; stride-8 costs $+44.4\%$ but runs $5.0\times$ faster. Near-term calibration (NLL@10) is essentially unaffected across strides; long-horizon calibration degrades more than the displacement numbers alone suggest, since NLL is a log-likelihood. The architecture thus degrades gracefully across temporal resolutions and is cheap to redeploy for a receiver's typical rate (full results and discussion in Appendix~\ref{app:stride}, Table~\ref{tab:stride}).

\begin{table}[hbt!]
\caption{Inference cost and monitoring capacity on one NVIDIA RTX 4000 Ada (float32). Latency: one request (a single input sequence, batch size 1), $K$ samples drawn in parallel, 20 Euler steps (CFM) or 100 DDIM steps (DDPM). Throughput: best over concurrent requests, at $K{=}20$. Aircraft per GPU: throughput $\times$ 3\,s, the mean ADS-B update interval, i.e.\ aircraft whose forecast can be refreshed at every update. $\geq$: the card was not saturated at the largest concurrency tested (1024).}
\label{tab:throughput}
\centering
\small
\begin{tabular}{l l c c c c c}
\toprule
Model & Type & Params & \multicolumn{2}{c}{Latency (ms)} & Throughput & Aircraft \\
 & & & $K{=}20$ & $K{=}50$ & (aircraft/s) & per GPU \\
\cmidrule(lr){4-5}
\midrule
LSTM-Det & --- & 1.5\,M & $1.15$ & $1.15$ & $\geq100{,}379$ & $\geq301{,}137$ \\
CVAE & CVAE & 1.5\,M & $4.06$ & $4.10$ & $16{,}388$ & $49{,}164$ \\
Diffusion Tiny & DDPM & 1.5\,M & $119.68$ & $229.50$ & $10.3$ & $30$ \\
Flow Tiny & CFM & 1.5\,M & $23.76$ & $47.25$ & $51.3$ & $153$ \\
Flow Small & CFM & 7.1\,M & $66.95$ & $160.80$ & $15.8$ & $47$ \\
Diffusion Large & DDPM & 20.7\,M & $773.16$ & $1870.65$ & $1.4$ & $4$ \\
Flow Large & CFM & 20.7\,M & $157.07$ & $375.82$ & $6.8$ & $20$ \\
Flow Large, stride 2 & CFM & 20.7\,M & $83.57$ & $205.27$ & $13.6$ & $40$ \\
Flow Large, stride 4 & CFM & 20.7\,M & $49.85$ & $107.80$ & $25.9$ & $77$ \\
Flow Large, stride 8 & CFM & 20.7\,M & $31.55$ & $54.68$ & $54.7$ & $164$ \\
\bottomrule
\end{tabular}
\end{table}

Because one request already occupies the GPU, batching several aircraft into a single forward pass barely raises throughput for the Transformer models: one RTX 4000 Ada serves about 50 aircraft per second with Flow Tiny (one 20-sample forecast each), 7 with Flow Large and 1.4 with Diffusion Large (Table~\ref{tab:throughput}). At the $3$\,s mean ADS-B update interval, a single GPU therefore keeps the forecasts of roughly 150 aircraft current with Flow Tiny, or 20 with Flow Large, and independent requests can be spread over further GPUs as traffic grows.

\begin{figure}[hbt!]
\centering
\includegraphics[width=0.6\linewidth]{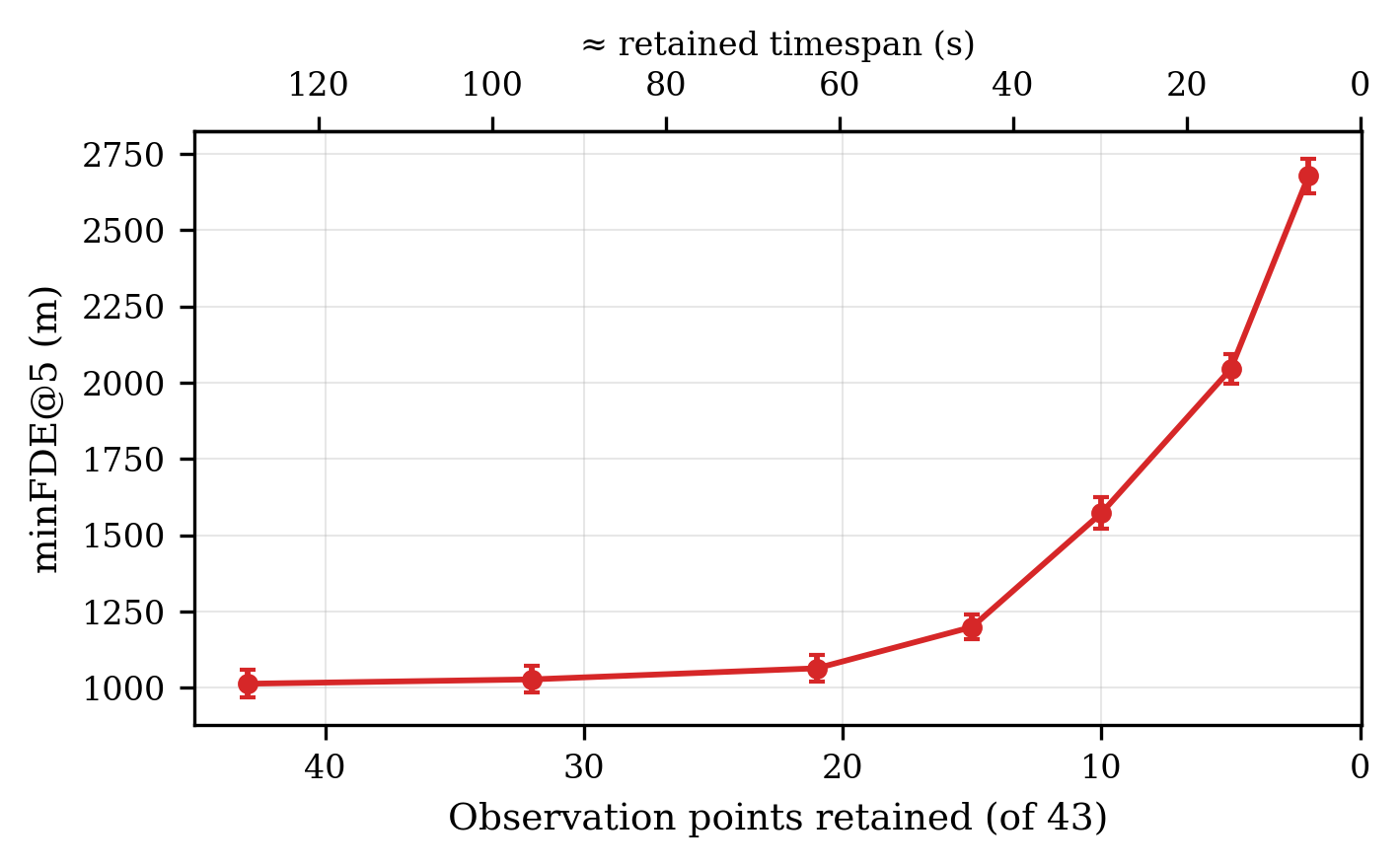}
\caption{Displacement error as a function of retained observation history. minFDE@5 (mean $\pm$ SEM, $N{=}2000$) as the observed prefix is shrunk from 43 points ($n_{\text{keep}}$) to progressively shorter spans; the earliest $43-n_{\text{keep}}$ tokens are frozen (repeated from the earliest retained sample) since the backbone's positional embeddings fix the input length at 43. Error is essentially flat down to $n_{\text{keep}}\approx21$ and rises sharply below $n_{\text{keep}}\approx15$.}
\label{fig:hist_length}
\end{figure}

The observed prefix can also be shortened at inference without retraining, by freezing the oldest tokens (repeated from the earliest retained observation) rather than truncating the fixed-length input. Figure~\ref{fig:hist_length} reports minFDE@5 as the retained history $n_{\text{keep}}$ shrinks from 43 to 2 points ($N{=}2000$ windows per level, $\approx\!1.4$ million forward passes in total): error stays essentially flat down to $n_{\text{keep}}\approx21$ (half the window, under $5\%$ cost) and rises sharply below $n_{\text{keep}}\approx15$, reaching $2.6\times$ the full-history error at $n_{\text{keep}}=2$. Because the frozen prefix is also mildly out-of-distribution relative to training data, this should be read as a conservative upper bound on the true cost of a genuinely shorter history.

\section{Spatial Bias}
\label{sec:spatial}

Without any chart or procedure supervision, FlowATC reproduces known Bay Area airspace structure such as approach turns, holding patterns, and descent profiles. This can be done by understanding the latent structures from the absolute Cartesian coordinates which encode geographic position. We examine this in three complementary ways: a catalogue of recurring turn geometries and their geographic relationship to published airspace procedures (A), qualitative evidence that the model's sampled trajectories reproduce them (B), and a quantitative comparison, at branch points where traffic divides, of the maneuvers FlowATC samples against those actually flown, including by aircraft that do not turn (C).

\subsection{Identifying Recurring Turn Patterns From ADS-B Trajectories}
\label{sec:turn_catalogue}

To evaluate whether the model reproduces location-specific maneuver structure, we use an independent catalogue of recurring turns built from ADS-B trajectories; all counts and evaluations below use its events from the collection of Section~\ref{sec:data}. A turn is defined as the transition between two locally stable ground-track segments. The catalogue is derived directly from observed trajectories and is not used during model training; the full detection, classification, and grouping protocol is given in Appendix~\ref{app:turn_catalogue}. Each turn is then associated geographically with published FAA navigation references using its detected start point. A turn with exactly one nearby reference within 1 nautical mile is assigned to the isolated-fix population, whereas a turn with multiple nearby references is assigned to the clustered/regional population. This association should be interpreted as spatial proximity rather than evidence that a particular fix, controller instruction, or published procedure caused the maneuver.

\begin{figure}[htbp]
    \centering
    \begin{subfigure}[t]{0.23\textwidth}
        \centering
        \includegraphics[width=\linewidth,height=4.9cm,keepaspectratio]{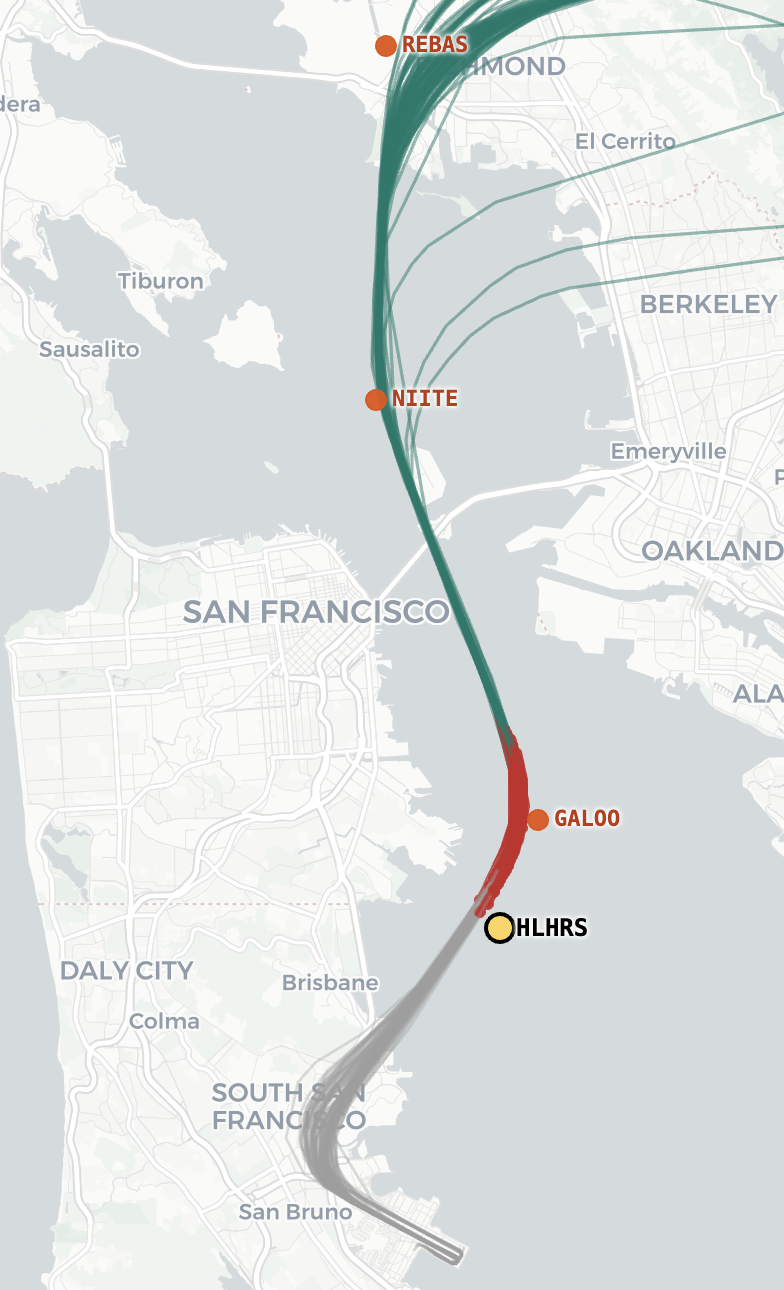}
        \caption{Turn pattern associated with the fix HLHRS.}
        \label{fig:niite_pattern_zoomed}
    \end{subfigure}
    \hfill
    \begin{subfigure}[t]{0.48\textwidth}
        \centering
        \includegraphics[width=\linewidth,height=4.9cm,keepaspectratio]{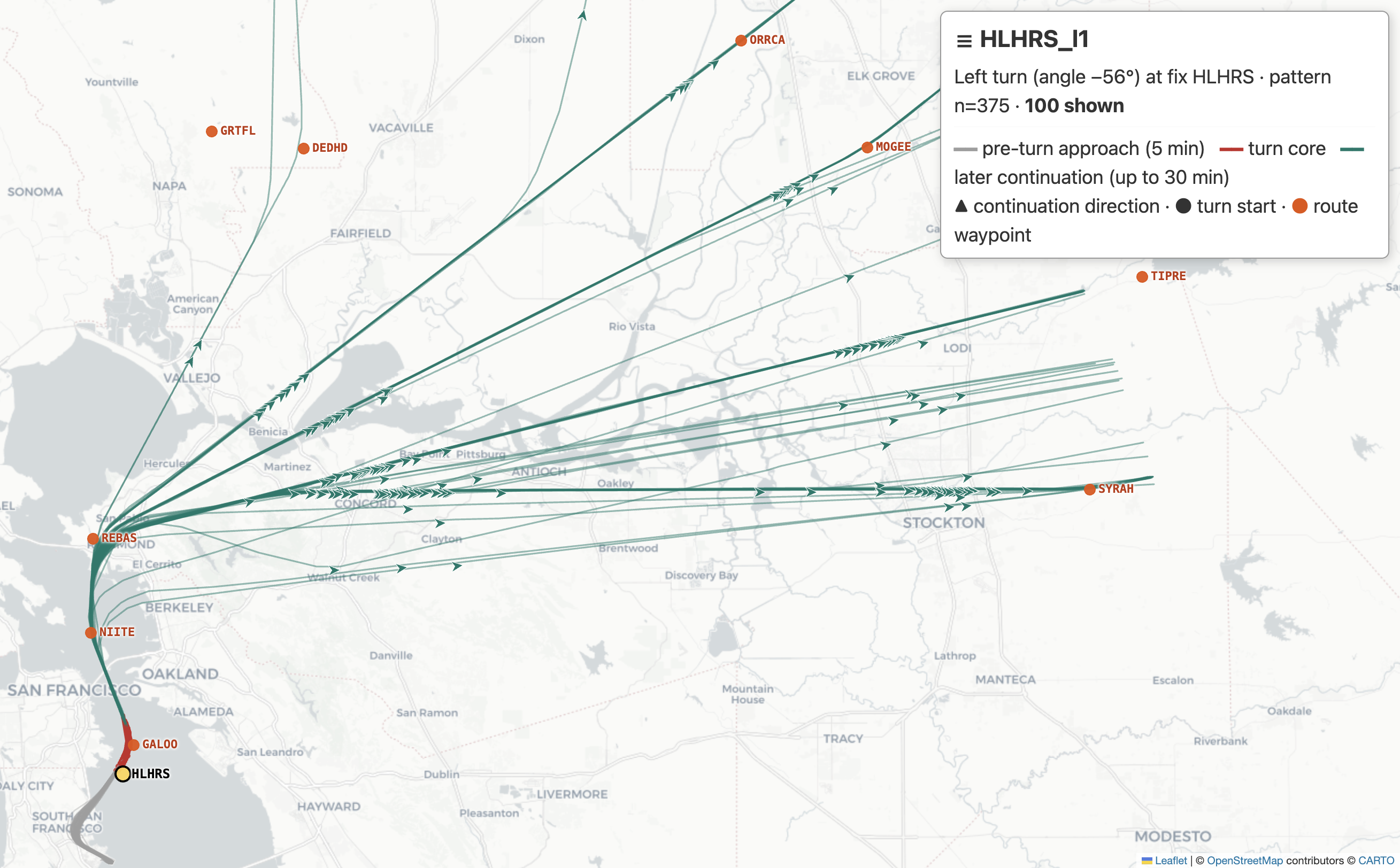}
        \caption{Full trajectory context of the recurring pattern.}
        \label{fig:niite_pattern_overview}
    \end{subfigure}
    \hfill
    \begin{subfigure}[t]{0.23\textwidth}
        \centering
        \includegraphics[width=\linewidth,height=4.9cm,keepaspectratio]{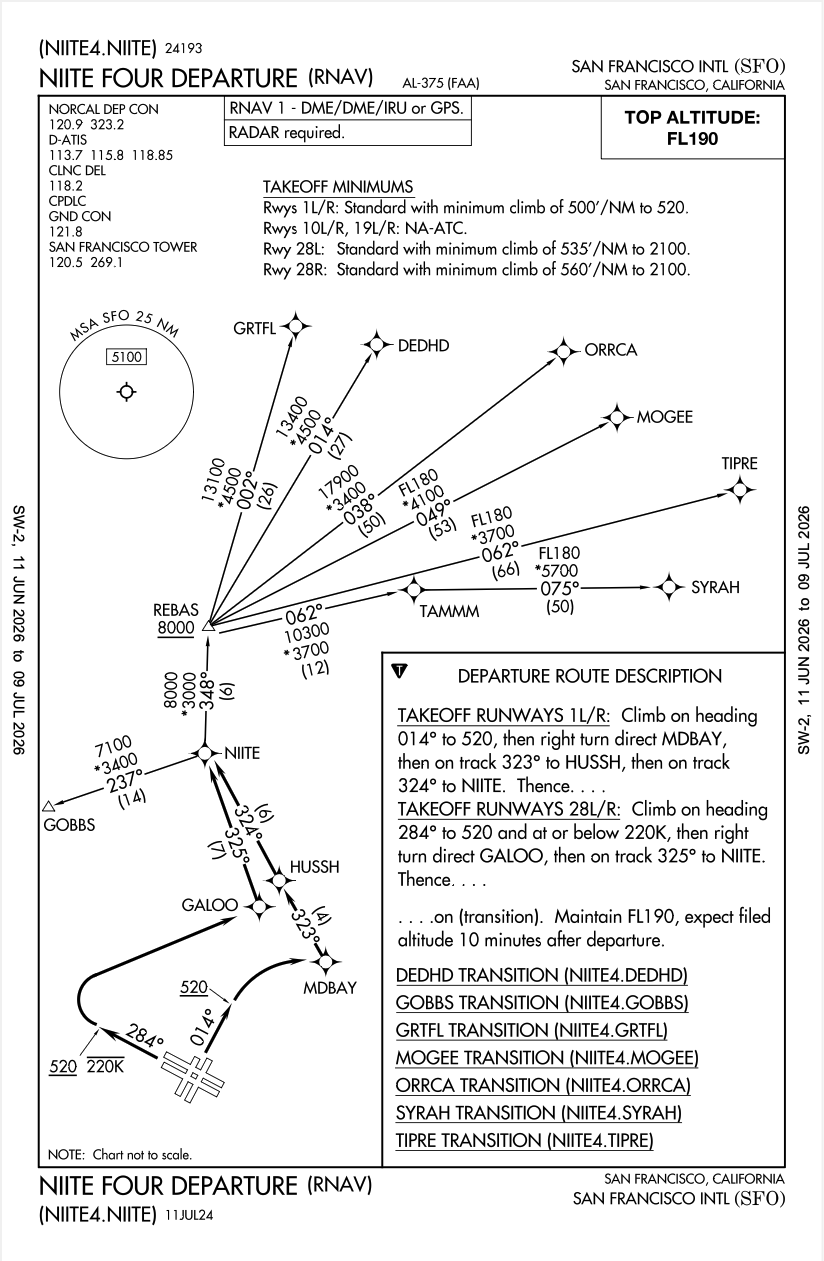}
        \caption{NIITE FOUR departure chart.}
        \label{fig:niite_four_chart}
    \end{subfigure}

    \caption{Recurring departure geometry recovered from ADS-B trajectories near SFO. \textbf{(a)} Turn pattern associated with the fix HLHRS. \textbf{(b)} Wider trajectory context of the recurring pattern. \textbf{(c)} The broad corridor and branching structure of the NIITE FOUR departure are visible in the extracted ADS-B pattern visualisations. Chart source: Federal Aviation Administration \cite{FAA2026NIITE4}}
    \label{fig:hlhrs_niite_procedure}
\end{figure}

The resulting catalog contains 433 recurring patterns and 32,765 non-overlapping turn events: 229 isolated-fix patterns comprising 12,076 events, and 204 clustered/regional patterns comprising 20,689 events. Each pattern is annotated with a pattern identifier, an associated fix or regional fix set, turn direction, central turn angle, empirical angular tolerance, qualitative compactness label, and event count. Compactness is reported as sharp, moderate, or broad. Beyond cataloguing frequent turns, grouping trajectories by recurring local geometry provides an empirical representation of the maneuver modes available at specific locations in the airspace. These modes often align with published procedure corridors, although geometry and fix proximity alone do not uniquely identify the procedure assigned to an individual flight.

Figure~\ref{fig:hlhrs_niite_procedure} illustrates this relationship for one recurring SFO departure pattern. The pattern is named \texttt{HLHRS\_l1}  because HLHRS is the only eligible navigation reference within the association radius of its detected turn start location. This name denotes geographic proximity and does not imply that HLHRS caused the maneuver or is part of the corresponding published procedure. Nevertheless, the observed trajectory corridor is broadly consistent with the NIITE FOUR departure shown alongside it.

The catalogue is designed to record recurring turns, so it contains only aircraft that turned. Evaluating a predictor requires the complementary view: every aircraft arriving at a decision point, whatever it does next. We therefore build on the catalogue to define branch points and collect all passages through them, recording the heading change actually flown, which is zero for aircraft that continue straight. At the 28 branch points retained for evaluation, a third of the 2{,}329 test passages fly a catalogued turn of that point and a third continue straight; the others fly maneuvers that the catalogue, restricted by design to recurring turns near navigation references, does not retain there.

\subsection{Qualitative Evidence of Learned Spatial Bias}
\label{sec:turn_qualitative}

The turn-pattern catalogue of Section~\ref{sec:turn_catalogue} provides a reference set of geographically recurring maneuvers against which model behavior can be inspected qualitatively. In this subsection, we compare extracted turn patterns with sampled futures from FlowATC in order to assess whether the model reproduces not only plausible trajectories, but also the spatially structured branching behavior observed in the ADS-B data.

\begin{figure}[htbp]
\centering
\begin{subfigure}[t]{0.48\linewidth}
\centering
\includegraphics[height=6.7cm]{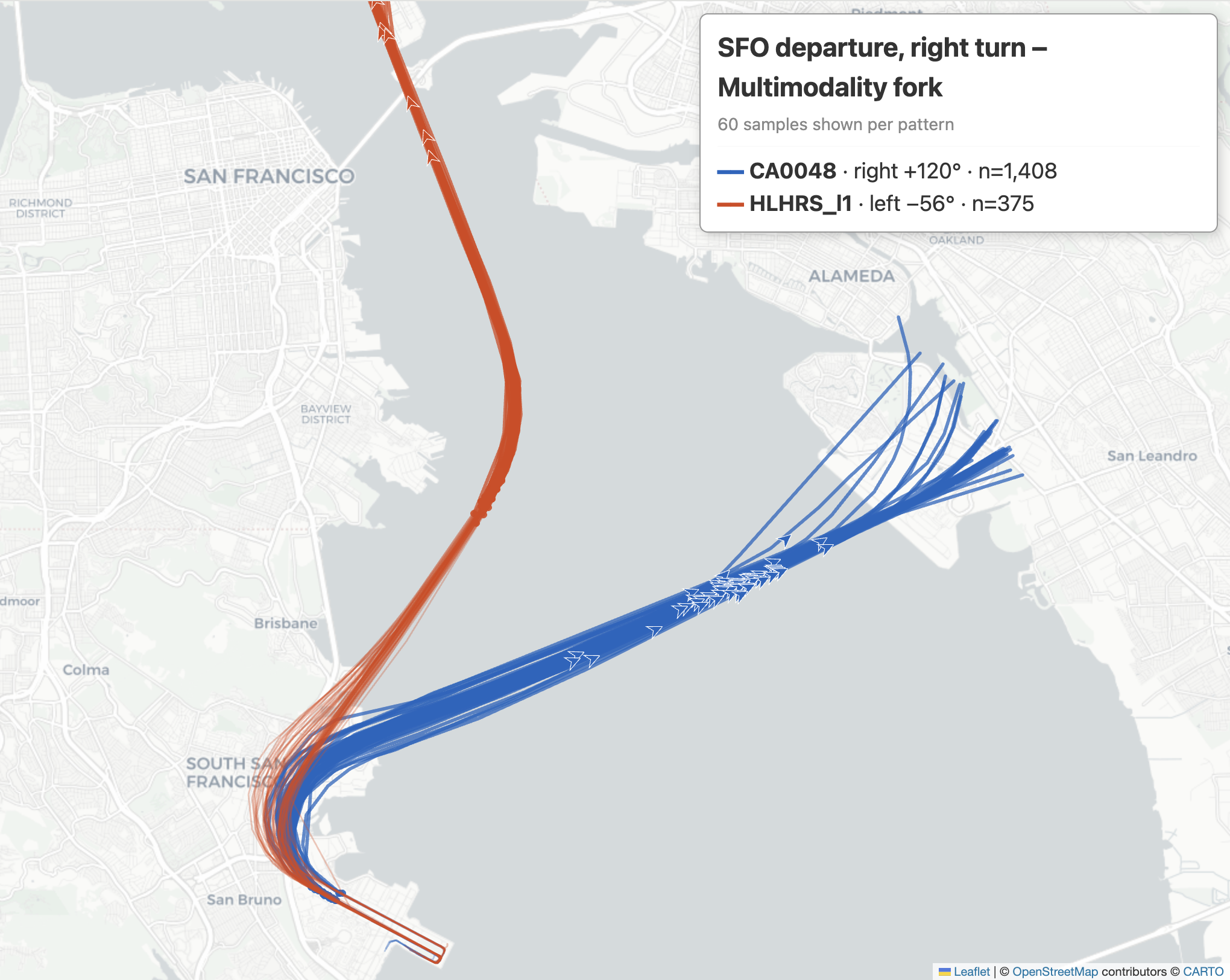}
\caption{Recurring SFO departure patterns \texttt{CA0048} and \texttt{HLHRS\_l1}.}
\label{fig:hlhrs_fork_reference}
\end{subfigure}
\hfill
\begin{subfigure}[t]{0.48\linewidth}
\centering
\includegraphics[height=6.7cm]{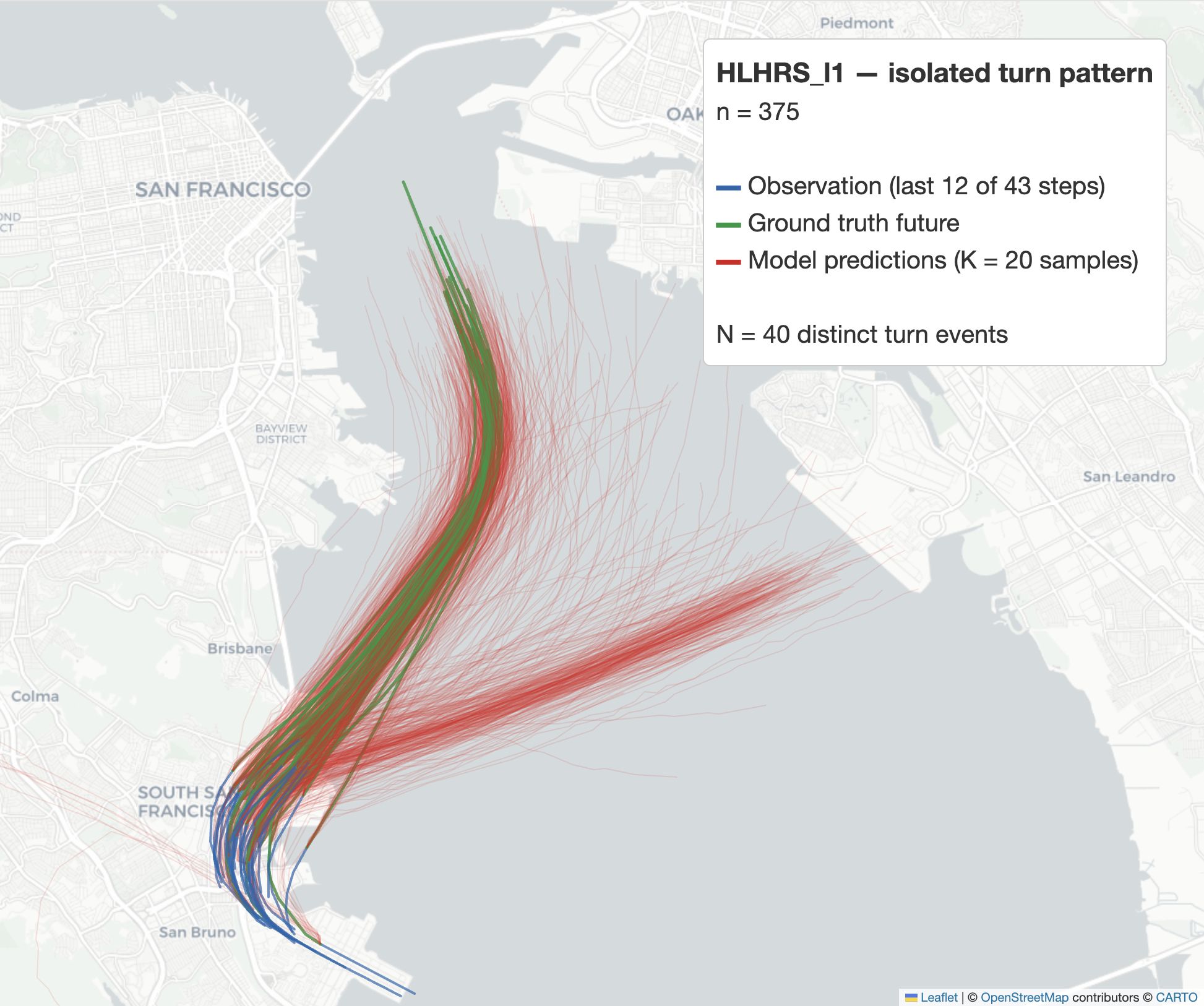}
\caption{Flow model samples for \texttt{HLHRS\_l1} test and validation events.}
\label{fig:hlhrs_fork_prediction}
\end{subfigure}
\caption{Multimodal departure structure near SFO.
\textbf{(a)} The recurring patterns \texttt{CA0048} ($n{=}1{,}408$) and \texttt{HLHRS\_l1} ($n{=}375$) share a similar initial right turn departure geometry before diverging. Lowercase \(n\) denotes the total number of occurrences in the turn pattern catalogue. The \texttt{CA0048} trajectories continue the turn toward an eastbound continuation, whereas \texttt{HLHRS\_l1} cuts the initial turn shorter, continues farther north, and subsequently forms the identified left turn pattern. \textbf{(b)} Predictions for $N{=}40$ distinct \texttt{HLHRS\_l1} turn events (15 test + 25 validation), with validation added because only 15 test aircraft yielded eligible prediction cases for this pattern. $K{=}20$ sampled futures per event. Blue shows the last 12 of the 43 observed points displayed for clarity, green the ground truth future, and red the model samples.}
\label{fig:hlhrs_multimodality}
\end{figure}

Figure~\ref{fig:hlhrs_multimodality} shows a representative two-mode example near San Francisco International Airport (SFO). The reference catalogue contains two departure patterns with similar initial geometry: clustered pattern \texttt{CA0048} (\(n=1{,}408\)) and \texttt{HLHRS\_l1} (\(n=375\)). Both depart SFO and initially execute a similar right turn, but their later geometries diverge. The \texttt{CA0048} trajectories maintain the turn for longer and continue predominantly eastward, whereas \texttt{HLHRS\_l1} cuts the initial turn shorter, continues farther north, and subsequently forms a left-turn pattern. The corresponding predictions are constructed from test and validation cases belonging only to \texttt{HLHRS\_l1}. Even in this setting, FlowATC samples populate both the northbound and eastbound continuations before the trajectories become fully distinguishable, indicating that FlowATC preserves multimodal uncertainty when the observed history is still compatible with more than one learned continuation. At the same time, the ground-truth future remains covered by a substantial subset of the samples.

\begin{figure}[htbp]
    \centering

    \begin{subfigure}[t]{0.32\textwidth}
        \centering
        \includegraphics[height=4.7cm]{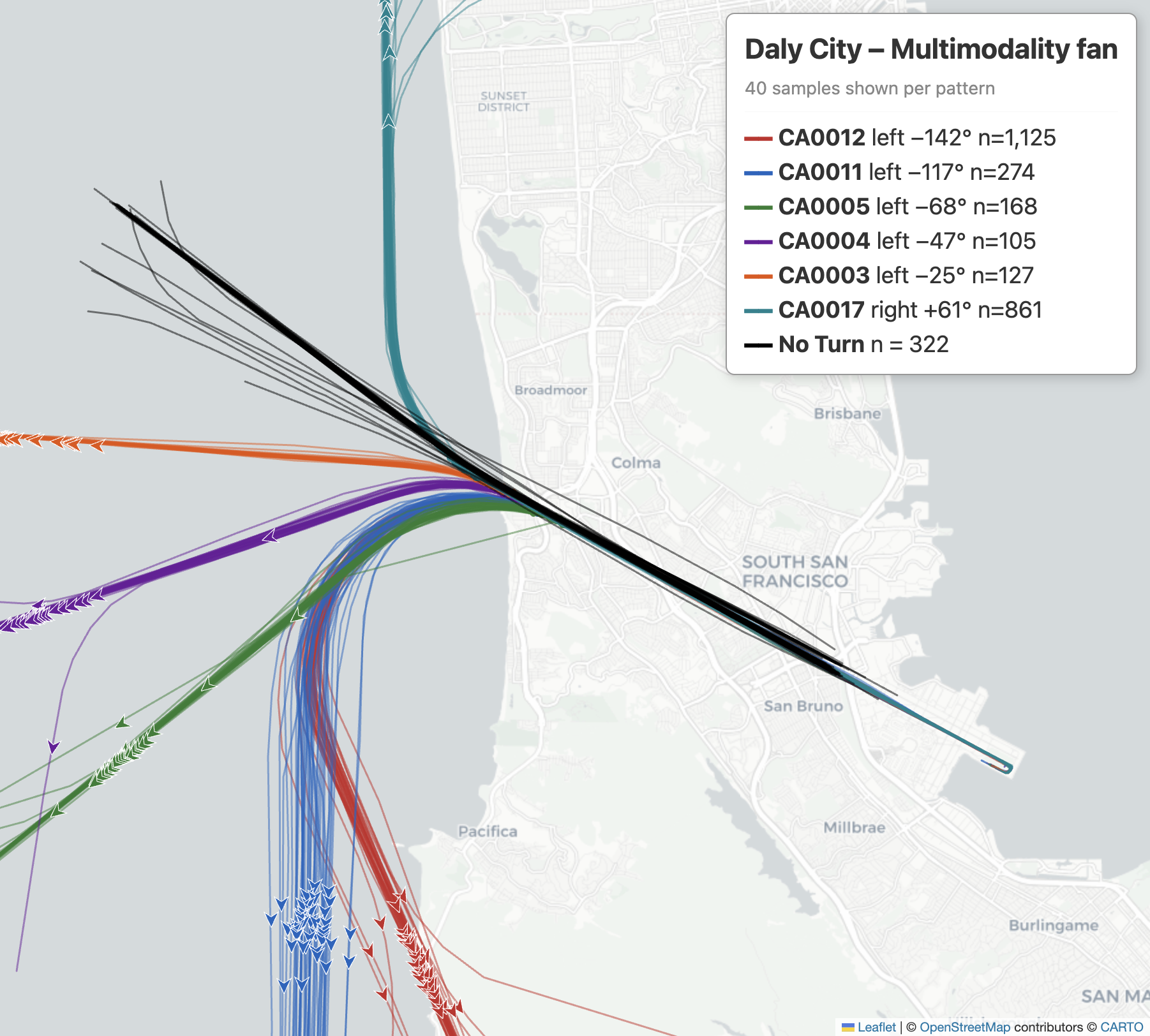}
        \caption{Six recurring catalogue patterns.}
        \label{fig:daly_fan_reference}
    \end{subfigure}
    \hfill
    \begin{subfigure}[t]{0.32\textwidth}
        \centering
        \includegraphics[height=4.7cm]{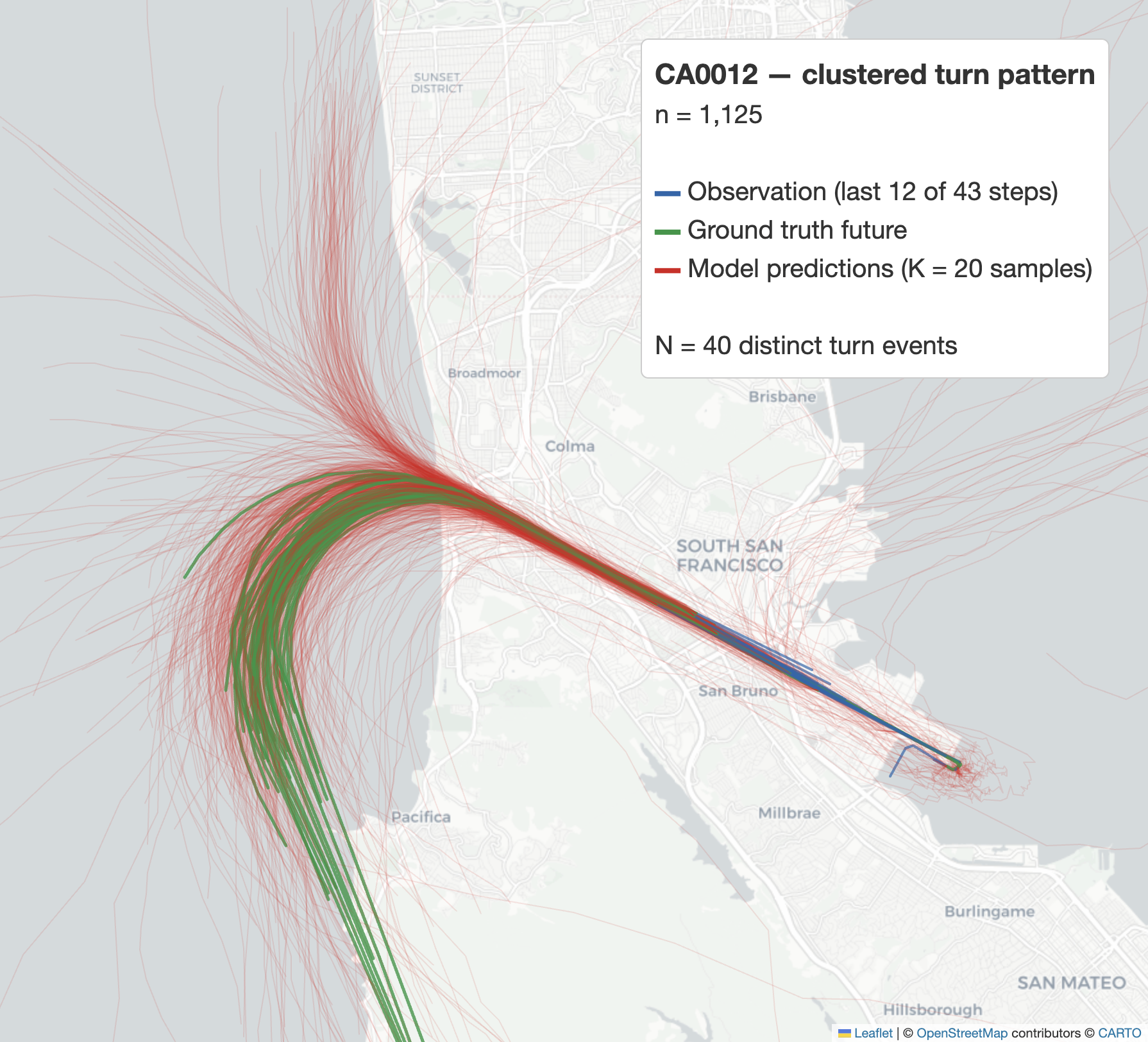}
        \caption{Flow-model samples for \texttt{CA0012} test events.}
        \label{fig:daly_fan_prediction_ca0012}
    \end{subfigure}
    \hfill
    \begin{subfigure}[t]{0.32\textwidth}
        \centering
        \includegraphics[height=4.7cm]{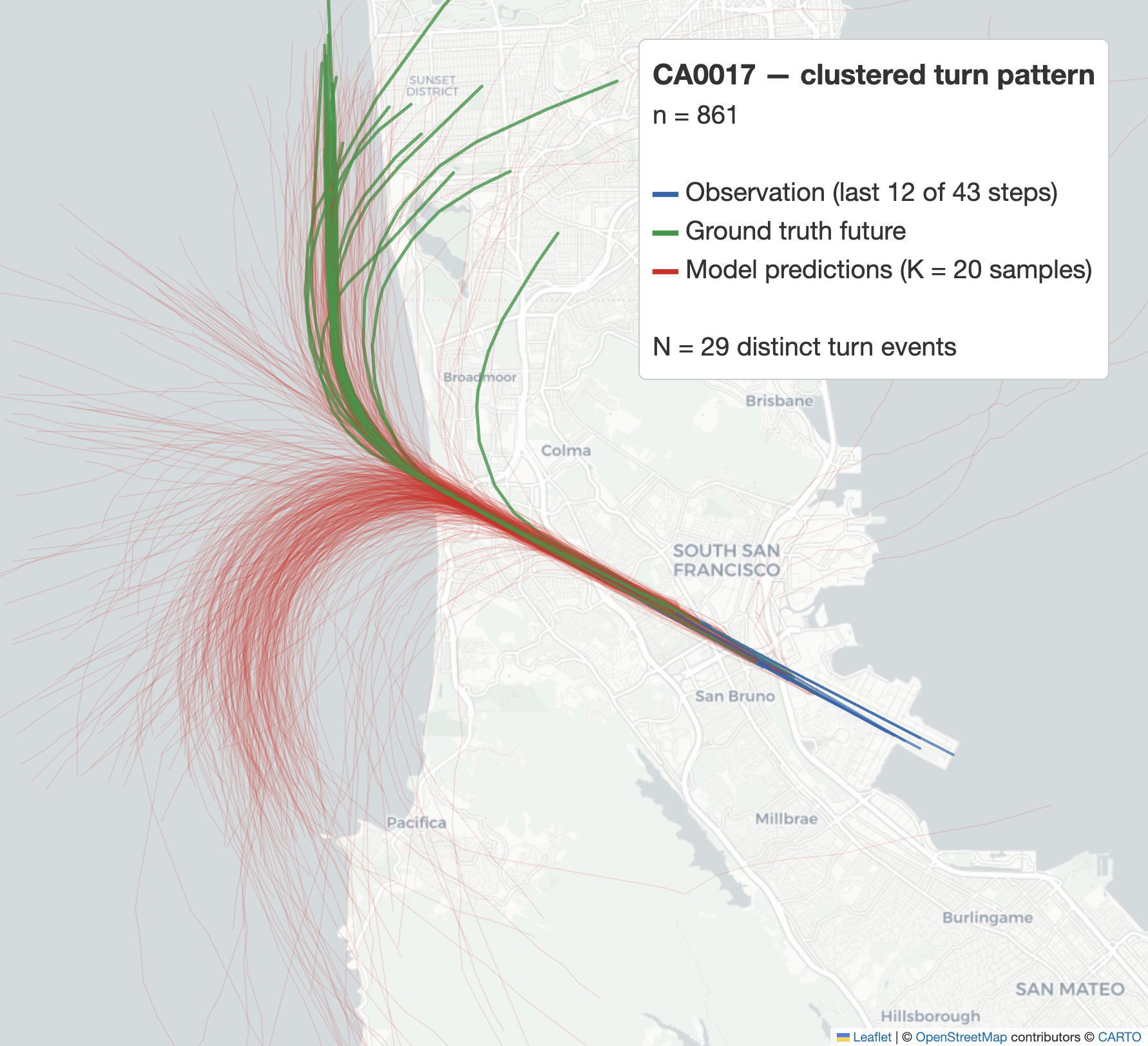}
        \caption{Flow-model samples for \texttt{CA0017} test events.}
        \label{fig:daly_fan_prediction_ca0017}
    \end{subfigure}

    \caption{A fan-shaped multimodal branching region near Daly City. \textbf{(a)} Six recurring catalogue patterns and a derived \emph{No Turn} cohort share a similar incoming trajectory before diverging into distinct continuations:
    \texttt{CA0012} (\(n=1{,}125\)),
    \texttt{CA0011} (\(n=274\)),
    \texttt{CA0005} (\(n=168\)),
    \texttt{CA0004} (\(n=105\)),
    \texttt{CA0003} (\(n=127\)),
    \texttt{CA0017} (\(n=861\)), and
    \emph{No Turn} (\(n=322\)).
    Lowercase \(n\) denotes the total number of occurrences in the turn pattern catalogue. The \emph{No Turn} group is a visualization-only cohort of historical flights that traverse the shared incoming corridor and continue through the branching region without entering one of the detected turn patterns. \textbf{(b)--(c)} Flow-model predictions for two separate test-only subsets. Panel \textbf{(b)} shows \(N=40\) distinct \texttt{CA0012} turn events, and panel \textbf{(c)} shows \(N=29\) distinct \texttt{CA0017} turn events, with \(K=20\) sampled futures per event. Blue shows the last 12 of the 43 observed points displayed for clarity, green the ground-truth future, and red the model samples.}
    \label{fig:fan_multimodality}
\end{figure}

A more complex example is shown in Fig.~\ref{fig:fan_multimodality}, where six recurring turn patterns and a derived No Turn cohort share a similar inbound segment and then diverge into a fan-shaped set of continuations near Daly City. The catalogue identifies the following turn patterns in this region: \texttt{CA0012} (\(n=1{,}125\)), \texttt{CA0011} (\(n=274\)), \texttt{CA0005} (\(n=168\)), \texttt{CA0004} (\(n=105\)), \texttt{CA0003} (\(n=127\)), and \texttt{CA0017} (\(n=861\)). In addition, the visualization includes a No Turn cohort (\(n=322\)), comprising historical flights that share the same incoming corridor and continue through the branching region without entering one of the detected turn patterns. The associated prediction panels show two separate test subsets, built from \(N=40\) (\texttt{CA0012}) and \(N=29\) (\texttt{CA0017}) distinct turn events with \(K=20\) sampled futures per event. In both cases, conditioned on the shared incoming geometry, FlowATC samples across several plausible outgoing branches while still covering the ground-truth continuation. However, not all branches are populated equally. The predictions concentrate most strongly on the historically dominant continuations, which are the patterns CA0012, CA0017 and straight continuation. This suggests that the observed prefix does not fully disambiguate the downstream branch, so FlowATC spreads probability mass over the modes roughly in proportion to how often each is flown (Section~\ref{sec:turnmode_quant}). Supporting this interpretation, the branches overlap substantially in turn-start altitude and speed, indicating that the preference for some continuations over others is not cleanly explained by simple kinematic separation. Overall, this behavior suggests that FlowATC has learned that the relevant uncertainty is not purely pointwise, but is organized around a discrete set of geographically recurring maneuver modes.

Taken together, these examples provide qualitative evidence that FlowATC learns location-specific airspace structure from trajectory data alone. Rather than collapsing toward an average future, it places samples across multiple historically observed maneuver modes when the observed history remains ambiguous. These qualitative observations motivate the quantitative evaluation at branch points of Section~\ref{sec:turnmode_quant}.

\subsection{Probabilistic Reproduction of maneuvers at Branch Points}
\label{sec:turnmode_quant}

Displacement error and KDE-NLL measure where probability mass lands in space, but neither asks whether the model reproduces the maneuver actually flown: at a given decision point, how often aircraft turn, how sharply, and with what spread. The branch passages of Section~\ref{sec:turn_catalogue} provide the reference needed to ask this directly, and, unlike the cataloged events, they include the aircraft that do not turn.

A branch point merges cataloged patterns lying within 1.5\,NM of each other whose incoming headings differ by at most $30^\circ$. A passage is selected from the observed history only: at the last observed point, the aircraft heads within $30^\circ$ of the branch point's incoming heading, points at it, would reach it within 15 to 90\,s at its current speed, and flies within the altitude band of the aircraft that fly its cataloged turns. One window is kept per passage. The altitude condition removes cruise overflights far above the procedure, which would otherwise inflate the share of straight flight. The outcome $\theta$ is the net heading change, unwrapped along the path, from the end of the observation until the path leaves a 4\,NM disc around the branch point. The same function is applied to the observed future and to each of the $K{=}50$ completions drawn by FlowATC, so the model and its reference are measured on the same footing. On passages that fly a cataloged turn, $\theta$ recovers the cataloged angle to a median error of $1.0^\circ$ without using the catalog's turn timing. We evaluate the 28 branch points that have at least 40 such passages in the held-out test split, 2{,}329 passages in total; the full protocol is given in Appendix~\ref{app:branch_protocol}.

\begin{figure}[hbt!]
\centering
\includegraphics[width=\linewidth]{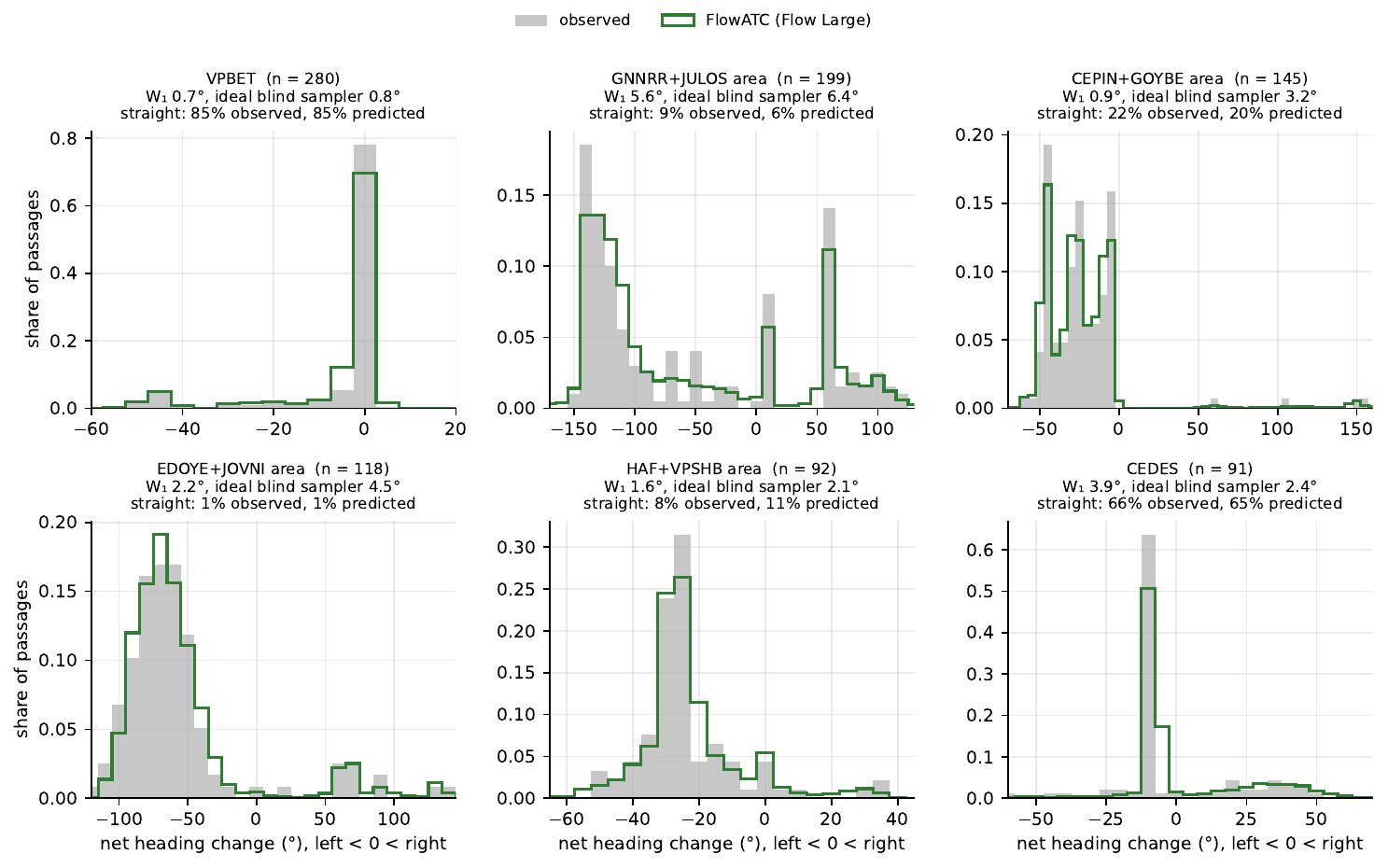}
\caption{Observed and sampled maneuver distributions at the six branch points with the most test passages. Gray bars: share of observed test passages per bin of net heading change $\theta$ (left turns negative, right turns positive, straight flight near zero). Green line: share of FlowATC samples (Flow Large, $K{=}50$ per passage) in the same bins. Titles give the number of passages $n$, the 1-Wasserstein distance $W_1$ between the two distributions, the value an ideal history-blind sampler would reach on this test set, and the observed and predicted shares of straight flight ($|\theta|<10^\circ$). The horizontal range covers the 1st to 99th percentiles of both distributions. No catalogue information enters the figure.}
\label{fig:branch_marginals}
\end{figure}

We first ask whether FlowATC reproduces the distribution of maneuvers flown at each branch point, pooled over all aircraft that reach it. Figure~\ref{fig:branch_marginals} overlays the observed and sampled distributions of $\theta$ at the six branch points with the most test passages; all 28 are shown in Appendix~\ref{app:branch_protocol}, Fig.~\ref{fig:branch_marginals_all}. FlowATC places its mass on the same modes, at the same angles and with comparable weights, including the three-branch fan near Daly City, and its share of straight flight matches the observed one within a few percentage points in every panel. Across the 28 branch points, $W_1$ between the pooled sampled and observed outcomes averages $3.3^\circ$, weighted by passages. This distance cannot vanish on a finite test set: an ideal sampler drawing from the true distribution of each branch point would itself score about $3.8^\circ$. FlowATC reaches that level at 16 of the 28 branch points, and predicts a straight-through share of $32.2\%$ against $32.9\%$ observed, with a correlation of $0.99$ across branch points.

Reproducing the pooled distribution does not require using the history: a sampler that ignored each aircraft and drew from the distribution of its branch point would do so as well. We therefore score each passage individually with the fair ensemble continuous ranked probability score (CRPS) \cite{ferro2014fair}, a fair finite-ensemble score that compares the $K$ sampled outcomes of one passage with the outcome actually flown, rewarding both accuracy and appropriate spread; for a single deterministic prediction it reduces to the absolute error. The reference is a history-blind sampler that, for each passage, draws from the observed outcomes of all other passages at the same branch point: it knows the exact test distribution but nothing about the aircraft. FlowATC reduces its CRPS by $47\%$ (95\% bootstrap interval over branch points: 36 to $62\%$), is better on $82\%$ of passages, and improves on it at all 28 branch points (Table~\ref{tab:branch_fidelity}). Its sampled intervals are close to calibrated: the central 80\% interval contains the observed outcome for $81\%$ of passages and the central 95\% interval for $92\%$, so its tails are slightly too narrow. FlowATC thus reproduces maneuvers at two levels: marginally, as the distribution of what all aircraft reaching a branch point do, and conditionally, as a close to calibrated distributionfor each aircraft given its observed history. 

\begin{table}[hbt!]
\caption{maneuver fidelity at the 28 branch points ($N{=}2{,}329$ test passages, $K{=}50$ samples each). $W_1$: 1-Wasserstein distance between pooled sampled and observed outcomes, averaged over branch points weighted by passages. Straight: share of outcomes with $|\theta|<10^\circ$ (observed: $32.9\%$). CRPS skill: relative CRPS reduction against the history-blind sampler, with 95\% bootstrap interval over branch points. Coverage: share of observed outcomes inside the central 80\% and 95\% sample intervals.}
\label{tab:branch_fidelity}
\centering
\small
\begin{tabular}{l c c c c c c}
\toprule
Model & Params & $W_1$ (deg) & Straight (\%) & CRPS skill & Cov.\ 80\% & Cov.\ 95\% \\
\midrule
Flow Large & 20.7\,M & $\mathbf{3.3}$ & $32.2$ & $\mathbf{0.47}$ [0.36, 0.62] & $81.2$ & $91.9$ \\
Flow Small & 7.1\,M  & $3.7$ & $33.1$ & $0.41$ [0.30, 0.58] & $80.9$ & $93.3$ \\
Flow Tiny  & 1.5\,M  & $9.6$ & $37.3$ & $0.25$ [0.12, 0.44] & $84.4$ & $96.0$ \\
CVAE       & 1.5\,M  & $10.4$ & $31.4$ & $0.25$ [0.13, 0.44] & $74.0$ & $83.6$ \\
\midrule
History-blind sampler   & --- & $3.8$  & $32.9$ & $0$ & --- & --- \\
Catalogued angles only  & --- & $27.2$ & $0.0$  & $-0.92$ [$-1.81$, $-0.35$] & --- & --- \\
Always straight         & --- & $37.1$ & $100$  & $-1.35$ & --- & --- \\
\bottomrule
\end{tabular}
\end{table}

Table~\ref{tab:branch_fidelity} mirrors the ordering of Section~\ref{sec:results}. Flow Small is almost as good as Flow Large, with a CRPS skill of $41\%$ against $47\%$, consistent with the saturation observed in Section~\ref{sec:results-scaling}, whereas Flow Tiny and the CVAE reach only $25\%$ and roughly triple the marginal distance. The CVAE is also the only model whose intervals are clearly too narrow: its central 95\% interval covers $84\%$ of outcomes. Drawing from the cataloged turn angles alone does worse than the history-blind sampler, because, by design, it describes only the aircraft that turn.

\section{Conclusion}
\label{sec:conclusion}
We presented FlowATC, a flow-matching architecture for aircraft trajectory prediction that frames the task as sequence inpainting. By concatenating observed and noisy tokens in a single DiT conditioned through AdaLN, FlowATC samples from the conditional distribution of future trajectories given the observed past. Four findings stand out. First, best-of-$K$ accuracy and distributional calibration are related but distinct properties, and generative models dominate deterministic and kinematic baselines primarily on the former: at matched capacity, CFM and DDPM both surpass the CVAE baseline by $31$--$41\%$ on minADE@20, while a single sample from either is no better, and sometimes worse, than a deterministic point estimate. $K$ independent samples are what turn this single-sample weakness into a calibrated spatial density over future positions that can serve as an input to downstream conflict-risk estimation. 
Second, CFM is consistently the strongest objective at matched parameter count, outperforming DDPM by $11$--$26\%$ in minADE@20 across the model sizes we test, though the gap narrows as both objectives scale, with DDPM's steeper scaling curve suggesting it may close entirely beyond the capacities evaluated here.
Third, FlowATC is operationally flexible along several axes: error degrades gracefully within the training horizon, temporal decimation to stride\,2 costs under $5\%$ in minADE@20 while roughly halving inference time, and stride\,8 gives a $5\times$ speedup at a $44\%$ accuracy cost. Queried beyond the training horizon, FlowATC remains the most accurate model up to 360\,s once several samples are drawn, and its calibration stays stable where the CVAE's diverges.

Lastly, FlowATC learns airspace structure without chart or procedure supervision: approach turns, holding patterns, and descent profiles emerge from data alone. At 28 branch points where traffic divides, it reproduces the distribution of maneuvers actually flown, including the third of aircraft that continue straight, and gives each aircraft a close to calibrated distribution that improves on a history-blind sampler by $47\%$ in CRPS.

\subsection*{Future Work}

FlowATC, in its current single-modality form, is designed as the first stage of a multi-modal predictor. Because the past enters only as additional context tokens, three conditioning streams can be appended without changing the loss or the inpainting mechanism:
\begin{enumerate}
    \item \textit{ATC voice instructions}: radio transcripts from Bay Area towers carry controller intent at maneuver initiation points; aligning voice embeddings with the trajectory space~\cite{brusset2026at} should sharpen the posterior at turn points.
    \item \textit{Weather fields}: wind forecasts and SIGMET polygons tokenized as additional context;
    \item \textit{FAA airspace structure}: the named fixes and directed procedure edges (STARs, SIDs, airways, holding patterns) of the Bay Area encoded as a graph conditioning signal.
\end{enumerate}
Each modality is expected to reduce density variance precisely at decision points where a single observed history is consistent with multiple controller instructions. A further stream is the flight plan itself: planned trajectories could be inserted conditionally into the future tokens as additional information.

\section*{Funding Sources}
This work was funded by École Polytechnique. Mathurin Petit was hosted at CITRIS and the Banatao Institute, University of California, Berkeley, under the Visiting Scholar Researcher program.

\section*{Acknowledgments}

We thank Professor Trevor Darrell and Jiahui Lei from the Berkeley Artificial Intelligence Research Lab (BAIR) for their advice regarding trajectory-applied flow motion.
We would also like to thank Tom Davis and John Robinson from Crown Innovations LLC, Parimal Kopardekar and James Murphy of NASA Ames Research Center, and Dragos Margineantu from Boeing for their insightful discussions regarding next-generation airspace decision support tools.
We also thank Professor Hans-Ludger Dienel from Technische Universität Berlin for coordinating with UC Berkeley.

\bibliography{references}
\newpage

\appendix
\section*{Appendices}
\addcontentsline{toc}{section}{Appendices}
\renewcommand{\thesubsection}{\Alph{subsection}}

The appendices collect supporting material that complements the main article and are useful for reproduction.

\subsection{Random Sample of Flow-Large Predictions on the Test Set}
\label{app:trajectory_mosaic}

\begin{figure}[p]
\centering
\includegraphics[width=0.9\linewidth]{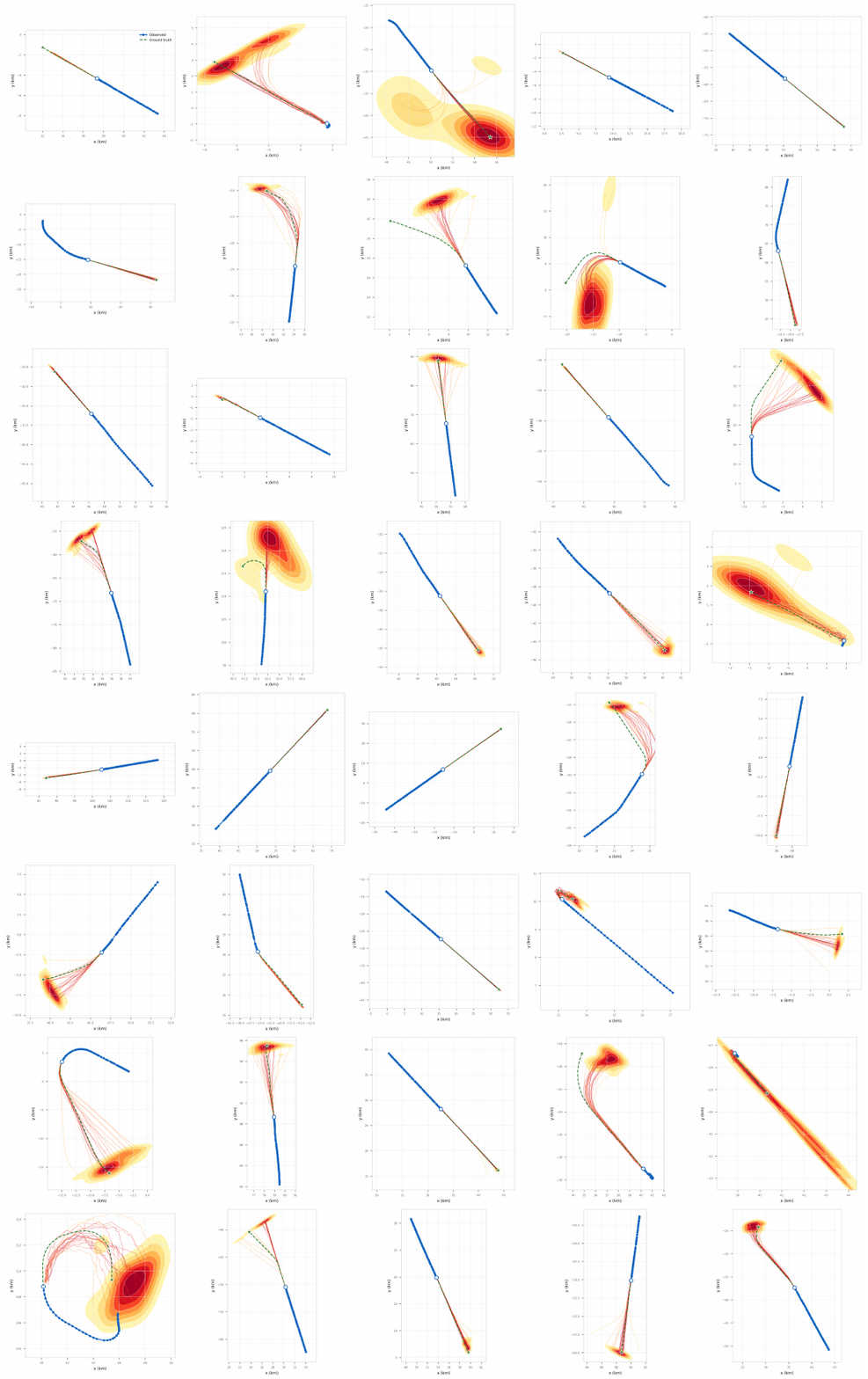}
\caption{Uncurated random sample of 40 test-set predictions}
\label{fig:trajectory_mosaic}
\end{figure}

To complement the curated multimodal examples of Section~\ref{sec:spatial}, which are deliberately selected to illustrate branching behavior at known decision points, Figure~\ref{fig:trajectory_mosaic} shows an \emph{uncurated} sample: 40 test windows drawn uniformly at random (fixed seed, no cherry-picking), predicted by Flow Large (20.7\,M parameters) at $K{=}20$ with the same 20-step Euler integration used throughout the paper. For each panel, blue marks the observed history (open circle: last observed point); the dashed green line and star mark the ground-truth future and its final position; the $K{=}20$ sampled trajectories are drawn as thin rays, coloured by the Gaussian KDE density (Scott's-rule bandwidth, Eq.~\eqref{eq:kde}) of their endpoint, with the same density shown as filled contours.

The large majority of panels show the predicted density tightly bracketing the ground truth, on both straight segments and turns, including one holding pattern reproduced almost exactly. A small minority (2 of the 40 panels shown) instead show the sampled density missing the realized outcome entirely. This is not hidden or excluded: an honest sample from a genuinely uncertain predictor occasionally misses, and this is precisely the behavior that KDE-NLL penalizes and that a best-of-$K$ figure alone would let a model hide.

\subsection{Data Preprocessing details}
\label{app:preprocessing}

Raw CSV files are ingested per aircraft, deduplicated on timestamp, and segmented into continuous flight segments: a new segment begins whenever the callsign changes or the inter-sample gap exceeds $120$\,s. Raw geodetic coordinates present three difficulties for neural processing: longitude is non-linear (one degree covers different physical distances at different latitudes), true track is a circular variable ($359^\circ\!\to\!0^\circ$), and position and velocity live at incompatible scales. We resolve these by projecting onto a local tangent-plane Cartesian frame centered on SFO ($\phi_0=37.6213^\circ$N, $\lambda_0=122.3790^\circ$W):
\begin{equation}
  x = (R+h)\cos\phi\,(\lambda-\lambda_0),\quad
  y = R\,(\phi-\phi_0),\quad
  z = h,
  \label{eq:cartesian}
\end{equation}
with velocity components $v_x=v\sin\theta$, $v_y=v\cos\theta$, $v_z=\dot h$.
\footnote{The altitude term $(R+h)$ appears in the east component $x$ but not in the north component $y$, where it would contribute a $<0.2\%$ correction ($h\ll R$); this asymmetry has no practical effect on the meter-scale features, which are z-score normalized regardless.} 
Segments are windowed into 86-point sequences with a stride of 10.

To expose the true timing to the network, we compute, for each sample, the elapsed time since the start of its window, apply a sinusoidal embedding ~\cite{vaswani2017attention} to it, and pass the result through a small MLP before adding it to the corresponding input token.  Flight segments and windows that are entirely on the ground (ADS-B \texttt{on\_ground} flag) are discarded during preprocessing, together with a small number of windows containing corrupted altitude readings ($<-50$\,m); every retained window contains at least one airborne sample. The train/validation/test split is drawn at the level of individual aircraft: the set of unique ICAO24 transponder codes is shuffled once with a fixed seed (42) and partitioned 85\%/10\%/5\%, and every window belonging to a given aircraft is assigned entirely to a single split, preventing correlated windows from leaking across splits.

\subsection{Flow matching and DDPM}
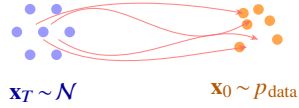
\begin{figure}[t!]
\centering
\begin{tikzpicture}[font=\small]
  \def\s{2.8}
  \begin{scope}
    \foreach \i in {1,...,6}{
      \pgfmathsetmacro{\ax}{0.35*cos(\i*60)}
      \pgfmathsetmacro{\ay}{0.35*sin(\i*60)}
      \fill[blue!40] (\ax,\ay) circle (2pt);
    }
    \fill[blue!40] (0,0) circle (2pt);
    \node[blue!60!black, below] at (0,-0.55) {$\mathbf{x}_0\!\sim\!\mathcal{N}$};
    \foreach \i/\dx/\dy in {1/0.2/0.15, 2/-0.15/0.25, 3/0.3/-0.1,
                            4/-0.2/-0.2, 5/0.05/0.3, 6/-0.1/0.1}{
      \fill[orange!70] (\s+\dx,\dy) circle (2pt);
    }
    \node[orange!70!black, below] at (\s,-0.55) {$\mathbf{x}_1\!\sim\! p_{\text{data}}$};
    \foreach \i/\ax/\ay/\bx/\by in {
        1/0.0/0.2/2.5/0.15,
        2/-0.15/0.0/2.7/0.25,
        3/0.2/-0.15/2.9/-0.1,
        4/0.0/-0.1/2.6/-0.2,
        5/0.1/0.3/2.65/0.3}{
      \draw[-{Stealth[length=3pt]},blue!50,opacity=0.7] (\ax,\ay) -- (\bx,\by);
    }
    \node[above] at (\s/2, 0.65) {\textbf{Flow Matching} (straight paths)};
  \end{scope}
  \begin{scope}[yshift=-2.2cm]
    \foreach \i in {1,...,6}{
      \pgfmathsetmacro{\ax}{0.35*cos(\i*60)}
      \pgfmathsetmacro{\ay}{0.35*sin(\i*60)}
      \fill[blue!40] (\ax,\ay) circle (2pt);
    }
    \fill[blue!40] (0,0) circle (2pt);
    \node[blue!60!black, below] at (0,-0.55) {$\mathbf{x}_T\!\sim\!\mathcal{N}$};
    \foreach \i/\dx/\dy in {1/0.2/0.15, 2/-0.15/0.25, 3/0.3/-0.1,
                            4/-0.2/-0.2, 5/0.05/0.3, 6/-0.1/0.1}{
      \fill[orange!70] (\s+\dx,\dy) circle (2pt);
    }
    \node[orange!70!black, below] at (\s,-0.55) {$\mathbf{x}_0\!\sim\! p_{\text{data}}$};
    \draw[-{Stealth[length=3pt]},red!60,opacity=0.8]
      (0.1,0.1) .. controls (1.0,0.8) and (1.6,-0.5) .. (2.75,0.15);
    \draw[-{Stealth[length=3pt]},red!60,opacity=0.8]
      (-0.1,0.1) .. controls (0.6,0.5) and (1.8,0.4) .. (2.65,0.25);
    \draw[-{Stealth[length=3pt]},red!60,opacity=0.8]
      (0.2,-0.1) .. controls (1.1,-0.7) and (2.0,0.2) .. (2.85,-0.1);
    \draw[-{Stealth[length=3pt]},red!60,opacity=0.8]
      (0.0,-0.1) .. controls (0.7,-0.6) and (1.9,-0.3) .. (2.7,-0.2);
    \node[above] at (\s/2, 0.65) {\textbf{DDPM} (curved / multi-step)};
  \end{scope}
\end{tikzpicture}
\caption{Conceptual comparison of Flow Matching and DDPM generation. Flow Matching transports samples along straight conditional paths, requiring fewer integration steps and yielding a simpler training objective. DDPM reverses a Markov noising chain along curved trajectories.}
\label{fig:cfm_concept}
\end{figure}

Diffusion models and flow matching are two families of generative models that learn to sample from a data distribution $p(\mathbf{x})$ by reversing a noise corruption process; both can be made \emph{conditional} by holding a context signal fixed while the target is denoised. Denoising Diffusion Probabilistic Models~\cite{ho2020ddpm} define a Markov chain that gradually adds Gaussian noise to a data sample over $T_{\text{diff}}$ steps until it is indistinguishable from pure noise. A network $\boldsymbol{\varepsilon}_\theta$ is trained to predict the added noise, enabling step-by-step denoising from a pure-noise sample at inference. The number of inference steps can be reduced (e.g.\ from 1000 to 100) with the DDIM sampler~\cite{song2021ddim} without significant quality loss. Conditional Flow Matching~\cite{lipman2022flow} learns a velocity field $u_\theta(\mathbf{x},t)$ that transports samples along straight-line paths from a Gaussian base distribution ($t=0$) to the data distribution ($t=1$). The objective is simple and deterministic: at any interpolation $\mathbf{x}_t=(1-t)\mathbf{x}_0+t\mathbf{x}_1$, the target velocity is the constant $(\mathbf{x}_1-\mathbf{x}_0)$. At inference, integrating the learned field with a few Euler steps produces a clean sample. Compared to DDPM, CFM has a simpler loss, requires fewer steps, and yields straighter sampling trajectories in data space (Fig.~\ref{fig:cfm_concept}).

\subsection{CVAE Decoder Variance at the First Prediction Step}
\label{app:cvae}

The CVAE's near-flat error curve at short horizon (Fig.~\ref{fig:degradation}a) and its NLL divergence at long horizon share a common cause. Decomposing the step-1 prediction on the held-out test set: the true one-step displacement averages 311\,m, the decoder's conditional-mean error is 111\,m (about twice the 56\,m of constant velocity), but the sampled Gaussian noise adds a further $\sigma_0 \approx 605$\,m horizontally. This $\sigma$ sits at its trained floor ($\approx$50\,m) for every subsequent step, so the excess variance at $n{=}1$ is a boundary effect rather than a genuine displacement error. Deterministic sampling (only latent draw, no decoder noise) confirms this: step-1 minADE@1 drops from 542.6\,m to 118.1\,m. The noise cannot simply be removed, however: because the prior places $\approx 95\%$ of its mass on a single category for a given observed history (12 of 25 categories carry non-negligible mass in aggregate, but conditionally the latent is nearly one-hot), disabling decoder noise collapses minADE@20 back toward minADE@1 ($853.4$ vs.\ $508.2$\,m) instead of the $K$ draws exploring $K$ distinct futures. 
The model's cumulative time encoding contributes to this: the decoder can recover the local inter-step gap for $n \geq 1$ by differencing two elapsed-time values it received itself, but the very first future gap (last observation to first prediction) is reachable only through the compressed history encoding. That defect, however, acts on the conditional mean, not on the decoder variance: at $n{=}1$ the mean is already within $111$\,m of the truth while $\sigma_0 \approx 605$\,m, so the offset is five times larger than any error the mean could contribute, and a better-informed mean cannot remove it. The best-of-$K$ comparisons that carry our headline results are driven by $\sigma$ and by the latent's conditional diversity, neither of which depends on this input, so we do not expect a time-corrected CVAE to change the $K{=}20$ ordering of Table~\ref{tab:model_comparison}.

The same limitation applies architecturally to the deterministic LSTM baseline, which uses an identical no-feedback decoder driven solely by cumulative $t_{\rm rel}$; its own step-1 error ($94.3$\,m) is nearly twice the steady per-step increment observed from step 2 onward ($\approx$50\,m), consistent with the same missing local time gap.

\subsection{Full $K\in\{1,5,20\}$ Displacement Sweep}
\label{app:degradation_full}

\begin{figure}[hbt!]
\centering
\includegraphics[width=\linewidth]{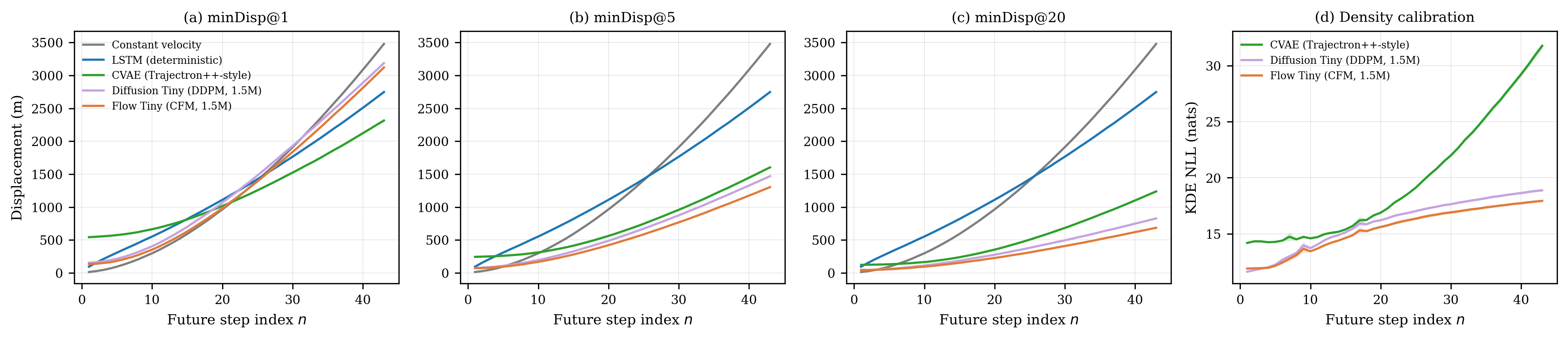}
\caption{Full $K\in\{1,5,20\}$ displacement-error sweep, extending
Fig.~\ref{fig:degradation} with the intermediate $K{=}5$ panel.}
\label{fig:degradation_full}
\end{figure}

Figure~\ref{fig:degradation}, in the main text, shows minDisp@1, minDisp@20 and density calibration only; $K{=}5$ is omitted there to keep the figure to three panels, since it sits between the $K{=}1$ and $K{=}20$ stories without adding a qualitatively new one. Figure~\ref{fig:degradation_full} reports the complete sweep for completeness.

\subsection{ADS-B Downsampling — Full Results}
\label{app:stride}

Because downsampling changes the physical time each future token represents, the KDE NLL columns of Table~\ref{tab:stride} are evaluated at the token index whose horizon is closest to the stride-1 reference rather than at a fixed token index (see table caption for the resulting horizons). Under this matched-horizon comparison, NLL@43 rises from $20.2$ to $22.2$--$23.0$\,nats across the three strides. Because NLL is a log-likelihood, this modest-looking increase in nats corresponds to the ground-truth density dropping by a factor of $e^{2.0}\!\approx\!7.4\times$ (stride-2) to $e^{2.8}\!\approx\!16.4\times$ (stride-4), a substantially larger effective loss of calibration than the nat values alone suggest. NLL@10, by contrast, is essentially unchanged and even improves slightly at stride-8 ($12.9$ vs.\ $13.3$\,nats, $e^{-0.4}\!\approx\!1.5\times$ \emph{more} likely). The architecture retrains effectively on lower-rate feeds, while the model's confidence at long horizons becomes markedly less trustworthy under coarser sampling, even though the displacement-error degradation (Table~\ref{tab:stride}) looks comparable in relative terms across horizons.

\begin{table}[hbt!]
\caption{Effect of temporal stride on Flow Large (20.7\,M params),
  $N=63{,}016$ test trajectories. Distances in metres~$\pm$\,SEM\@.
  KDE NLL in nats, reported at the token whose physical horizon is
  nearest the stride-1 reference ($n{=}10,20,43 \approx 30,60,129$\,s);
  stride models thus reach $\{30,60,132\}$\,s (stride 2),
  $\{36,60,132\}$\,s (stride 4) and $\{24,48,120\}$\,s (stride 8). Bold = reference.}
\label{tab:stride}
\centering
\footnotesize
\setlength{\tabcolsep}{4pt}
\resizebox{\textwidth}{!}{%
\begin{tabular}{l c cc cc cc ccc c}
\toprule
Model & Stride
  & \multicolumn{2}{c}{$K=1$}
  & \multicolumn{2}{c}{$K=5$}
  & \multicolumn{2}{c}{$K=20$}
  & \multicolumn{3}{c}{KDE NLL@$n$}
  & ms/pred \\
\cmidrule(lr){3-4}\cmidrule(lr){5-6}\cmidrule(lr){7-8}\cmidrule(lr){9-11}
 & & minADE & minFDE & minADE & minFDE & minADE & minFDE
   & $n{=}10$ & $n{=}20$ & $n{=}43$ & \\
\midrule
Flow Large & 1 & $\mathbf{903.4}_{\pm5.9}$  & $\mathbf{2156.8}_{\pm14.2}$ & $\mathbf{467.3}_{\pm3.4}$ & $\mathbf{1013.2}_{\pm7.8}$ & $\mathbf{307.4}_{\pm2.4}$ & $\mathbf{576.8}_{\pm5.3}$ & $\mathbf{13.3}_{\pm0.1}$ & $\mathbf{17.5}_{\pm0.5}$ & $\mathbf{20.2}_{\pm0.4}$ & 157.07 \\
Flow Large & 2 & $986.7_{\pm6.6}$  & $2315.9_{\pm15.9}$ & $494.2_{\pm3.5}$  & $1044.4_{\pm8.1}$  & $322.6_{\pm2.3}$  & $585.5_{\pm5.1}$  & $13.8_{\pm0.1}$ & $18.2_{\pm0.4}$ & $22.2_{\pm0.9}$ & 83.57 \\
Flow Large & 4 & $1031.2_{\pm6.7}$ & $2271.0_{\pm15.1}$ & $531.2_{\pm3.8}$  & $1066.3_{\pm8.3}$  & $352.5_{\pm2.7}$  & $619.8_{\pm5.6}$  & $15.1_{\pm0.2}$ & $18.8_{\pm0.4}$ & $23.0_{\pm0.7}$ & 49.85 \\
Flow Large & 8 & $1310.7_{\pm8.5}$ & $2631.3_{\pm17.5}$ & $674.7_{\pm4.8}$  & $1239.2_{\pm9.7}$  & $443.8_{\pm3.2}$  & $711.3_{\pm6.3}$  & $12.9_{\pm0.1}$ & $16.2_{\pm0.2}$ & $22.5_{\pm0.6}$ & 31.55 \\
\bottomrule
\end{tabular}}
\end{table}

\subsection{Full Extrapolation Sweep and Shard-Stitching Detail}
\label{app:extrapolation_full}

\begin{figure}[hbt!]
\centering
\includegraphics[width=\linewidth]{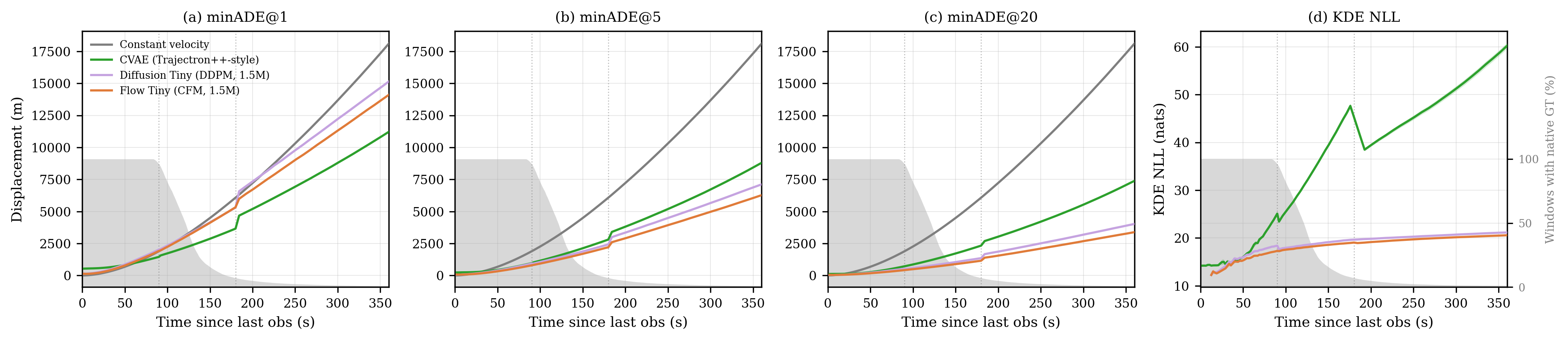}
\caption{Full $K\in\{1,5,20\}$ extrapolation sweep, extending Fig.~\ref{fig:extrapolation_error} with the intermediate $K{=}5$ panel.}
\label{fig:extrapolation_error_full}
\end{figure}

Figure~\ref{fig:extrapolation_error} in the main text omits minDisp@5; Figure~\ref{fig:extrapolation_error_full} reports the complete $K\in\{1,5,20\}$ sweep. Because $\Delta t_{\mathrm{virt}} = H_{\max}/43$ is fixed once $H_{\max}$ is chosen, a single run only gives good local temporal resolution near its own $H_{\max}$: a run with $H_{\max}=360$\,s has $\Delta t_{\mathrm{virt}}\approx8.4$\,s/token, under-resolving the $0$--$90$\,s range relative to a dedicated $H_{\max}=90$\,s run ($\Delta t_{\mathrm{virt}}\approx2.1$\,s/token). We therefore stitch three independent shards, each with its own 43-token resampling and its own $K{=}50$ posterior draws, keeping for each time band the shard with the finest resolution available: $(0,90]$\,s from $H_{\max}{=}90$, $(90,180]$\,s from $H_{\max}{=}180$, $(180,360]$\,s from $H_{\max}{=}360$. The visible seam at each band boundary (most noticeable at $t\approx180$\,s) is not a change in model behavior but a change of estimator: the two shards on either side differ simultaneously in token resolution, in the finite-sample statistics computed from a different draw of the $K$ posterior samples, and in which subset of test windows passes the coverage filter at that instant.

\subsection{Turn Catalogue: Detection, Classification, and Grouping Protocol}
\label{app:turn_catalogue}

We first retain sustained airborne trajectory segments by requiring a geometric altitude of at least $150\,\mathrm{m}$ and a ground speed of at least $25\,\mathrm{m\,s^{-1}}$. After filtering, gaps longer than $30\,\mathrm{s}$ split a track into separate segments, and segments containing fewer than five points are discarded.

\begin{figure}[hbt!]
\centering
\includegraphics[width=0.38\linewidth]{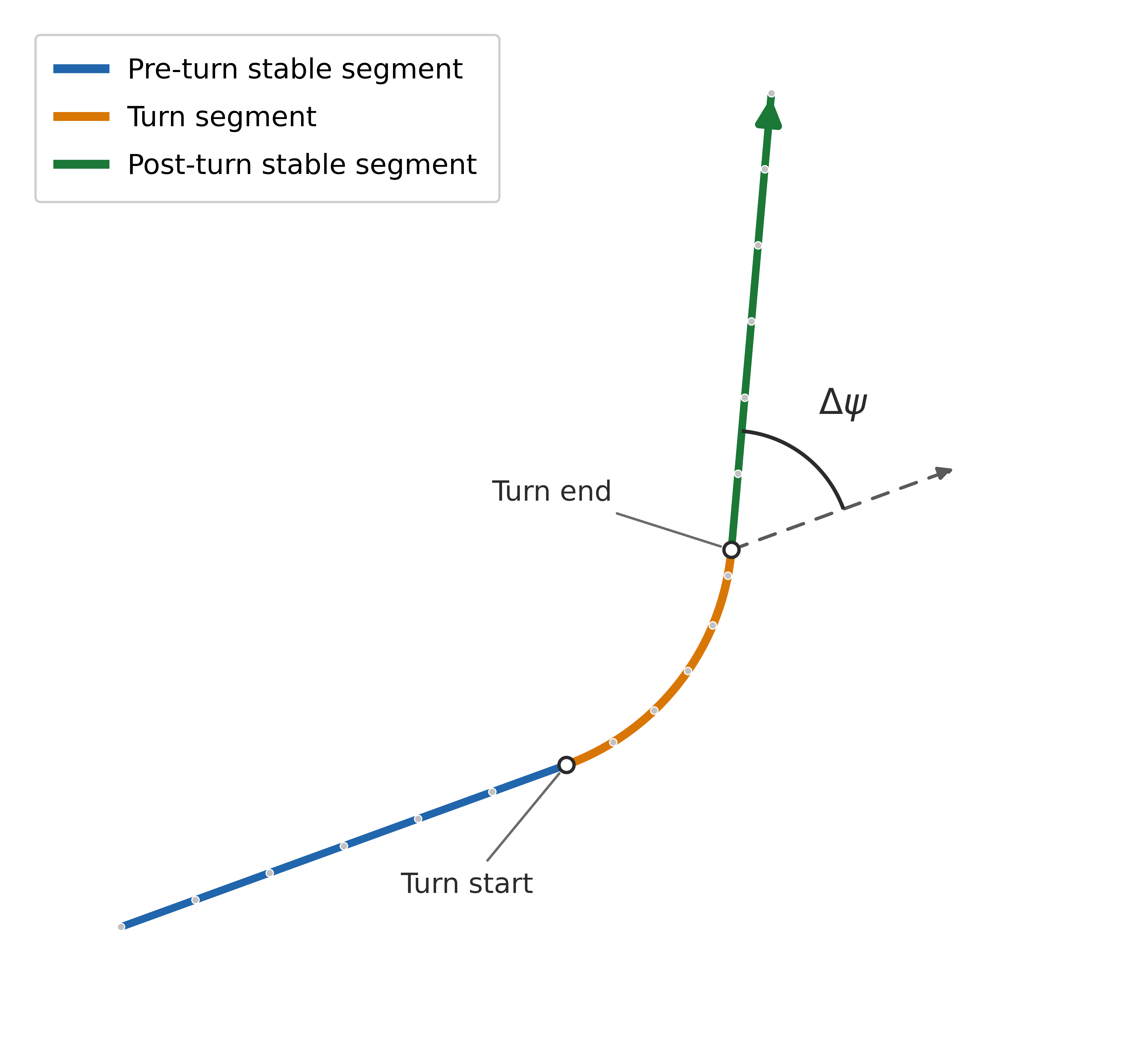}
\caption{A turn event is defined as the transition between two locally stable ground-track segments. The detected turn start and end delimit this transition, while the turn angle, $\Delta\psi$, is the wrapped angular difference between the mean ground-track directions of the pre-turn and post-turn stable segments.}
\label{fig:turn_detection_schematic}
\end{figure}

The detector uses the ADS-B-reported true track, smoothed with a short circular moving average to avoid discontinuities at $0^\circ/360^\circ$. A stable segment is a sequence over which the pointwise ground-track rate remains below $0.5^\circ\,\mathrm{s}^{-1}$ and the cumulative drift from the initial course remains below $6^\circ$. For two consecutive stable segments with circular-mean directions $\bar{\psi}_{\mathrm{pre}}$ and $\bar{\psi}_{\mathrm{post}}$, the signed turn angle is

\begin{equation}
\Delta\psi =
\left[
\left(
\bar{\psi}_{\mathrm{post}}
-
\bar{\psi}_{\mathrm{pre}}
+
180^\circ
\right)
\bmod 360^\circ
\right]
-
180^\circ .
\label{eq:wrapped_turn_angle}
\end{equation}

Positive and negative values denote right and left turns, respectively, and the interval between the two stable segments defines the detected turn start and end. Changes of at least $20^\circ$ are retained directly as major turns; smaller changes between $10^\circ$ and $20^\circ$ are retained only when the observed path geometry is consistent with the reported ground-track change. Smaller course adjustments are excluded from the pattern catalogue. Extending the association rule of Section~\ref{sec:turn_catalogue}, turns without an eligible navigation reference within 1 nautical mile are left unmatched and excluded from both pattern populations. After detecting individual turn events, we group events that occur in the same local area and exhibit similar incoming and outgoing ground-track directions, together with a similar overall signed ground-track change, $\Delta\psi$. When a turn start is associated with a single navigation reference, patterns are formed at the fix level; in dense terminal regions with several nearby references, turns are grouped regionally. Only recurring and geometrically coherent groups are retained as catalogue patterns, with limited manual review used to resolve evident duplicates or inconsistent cases.

\subsection{Branch Points, Passages and Scores}
\label{app:branch_protocol}

\paragraph{Branch points.} Catalogued patterns with at least 60 events are located at the position of their associated fix (isolated family) or at the median start of their turns (clustered family). Patterns lying within 1.5\,NM of each other whose incoming headings differ by at most $30^\circ$ are merged; the branch point's incoming heading is the event-weighted circular mean. This yields 80 branch points.

\paragraph{Passages.} A window is a candidate passage if, at its last observed point, the aircraft is airborne (altitude at least 150\,m, ground speed at least 25\,m\,s$^{-1}$, as in the catalogue), heads within $30^\circ$ of the incoming heading, has the branch point within $25^\circ$ of its heading, and would reach it in 15 to 90\,s at its current speed. Candidate windows of the same aircraft separated by less than 900\,s form one passage, represented by the window whose time to the branch point is closest to 25\,s. The altitude band of a branch point spans the 5th to 95th percentile, widened by 500\,m, of the altitudes of training and validation passages that fly one of its catalogued turns; branch points with fewer than 10 such passages have no band. Test passages outside the band are discarded. The evaluation uses the 28 branch points with at least 40 remaining test passages.

\paragraph{Outcome.} Headings are computed over four-step segments and unwrapped along the path, starting from the last observed heading. The outcome $\theta$ is the accumulated heading change at the first point, after the closest approach, where the path is more than 4\,NM from the branch point. When the path is still within the disc at the end of the window ($26.9\%$ of test passages) or never enters it ($0.6\%$), $\theta$ is taken at the end of the window; the same rule applies to every sample. On the 771 test passages that fly a catalogued turn of their branch point, $\theta$ matches the catalogued angle to a median of $1.0^\circ$, and to within $15^\circ$ for $86\%$ of them.

\paragraph{Scores.} $W_1$ compares the $K n$ pooled sampled outcomes of a branch point with its $n$ observed outcomes. The level of an ideal history-blind sampler is estimated as half the mean $W_1$ between two random halves of the observed outcomes, since $W_1$ between $n$ empirical values and their law scales as $n^{-1/2}$; simulations on mixtures shaped like branch points agree within $4\%$. For a passage with samples $\theta^{(1)},\dots,\theta^{(K)}$ and observed outcome $\theta$, the ensemble CRPS is
\begin{equation}
  \mathrm{CRPS} = \frac{1}{K}\sum_{k} \bigl|\theta^{(k)}-\theta\bigr|
  - \frac{1}{2K(K-1)}\sum_{k\neq l}\bigl|\theta^{(k)}-\theta^{(l)}\bigr| .
\end{equation}
The history-blind sampler uses as ensemble the observed outcomes of all other test passages at the same branch point. Bootstrap intervals resample branch points 1000 times.

\begin{figure}[p]
\centering
\includegraphics[width=\linewidth]{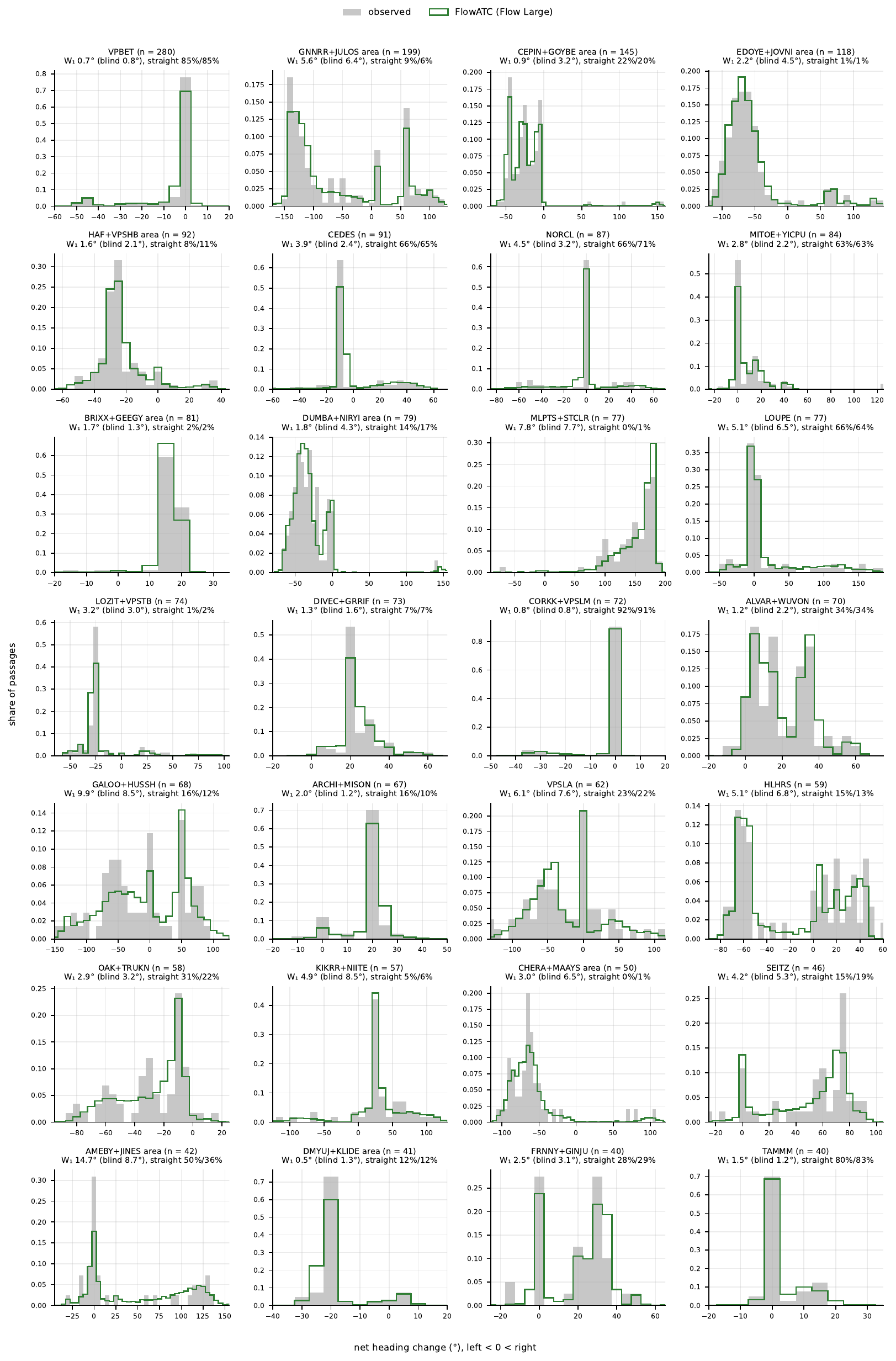}
\caption{Observed and sampled maneuver distributions at all 28 evaluated branch points, sorted by number of test passages; construction as in Fig.~\ref{fig:branch_marginals}. Titles give $W_1$, the ideal history-blind level, and the observed and predicted shares of straight flight.} 
\label{fig:branch_marginals_all}
\end{figure}

\end{document}